\documentclass[letterpaper, 10 pt, conference]{ieeeconf}  % Comment this line out if you need a4paper

\IEEEoverridecommandlockouts                              % This command is only needed if 
\usepackage{graphics} % for pdf, bitmapped graphics files
\usepackage{graphicx}
\usepackage{amsmath} % assumes amsmath package installed
\usepackage{xcolor}
\usepackage{multirow}
\usepackage{bm}
\usepackage{makecell}
\usepackage{url}
\usepackage{float}
\usepackage{subcaption}
\usepackage{booktabs,makecell,multirow}
\usepackage{amsfonts}
\usepackage{mathrsfs}

\usepackage[some]{background}
\SetBgScale{1}
\SetBgContents{\parbox{10cm}{%
  \Huge Draft:  \today\\[18cm]\rotatebox{180}{\Huge Draft:  \today}}}
\SetBgColor{red}
\SetBgAngle{270}
\SetBgOpacity{0.2}

\title{\LARGE \bf
Adaptive Admittance Control for Safety-Critical Physical Human Robot Collaboration}

\author{Yuzhu Sun, Mien Van, Stephen McIlvanna, Seán McLoone and Dariusz Ceglarek% <-this % stops a space
\thanks{Yuzhu Sun, Mien Van (corresponding author), Stephen McIlvanna and Seán McLoone are with the Centre for Intelligent Autonomous Manufacturing Systems, School of Electronics, Electrical Engineering and Computer Science, Queen's University Belfast, Northern Ireland, UK.
        (email: ysun32@qub.ac.uk; m.van@qub.ac.uk; smcilvanna01@qub.ac.uk; s.mcloone@qub.ac.uk)}%
\thanks{Dariusz Ceglarek is with the Warwick Manufacturing Group, University of Warwick, Coventry, UK.
        (email: d.j.ceglarek@warwick.ac.uk)}%
}

\begin{document}

\maketitle
\thispagestyle{empty}
\pagestyle{empty}

% Command to add the watermark in the title page specifying the version 
%\BgThispage
%%%%%%

%%%%%%%%%%%%%%%%%%%%%%%%%%%%%%%%%%%%%%%%%%%%%%%%%%%%%%%%%%%%%%%%%%%%%%%%%%%%%%%%
\begin{abstract}
Physical human-robot collaboration requires strict safety guarantees since robots and humans work in a shared workspace. This letter presents a novel control framework to handle safety-critical position-based constraints for human-robot physical interaction. The proposed methodology is based on admittance control, exponential control barrier functions (ECBFs) and quadratic program (QP) to achieve compliance during the force interaction between human and robot, while simultaneously guaranteeing safety constraints. In particular, the formulation of admittance control is rewritten as a second-order nonlinear control system, and the interaction forces between humans and robots are regarded as the control input. A virtual force feedback for admittance control is provided in real-time by using the ECBFs-QP framework as a compensator of the external human forces. A safe trajectory is therefore derived from the proposed adaptive admittance control scheme for a low-level controller to track. The innovation of the proposed approach is that the proposed controller will enable the robot to comply with human forces with natural fluidity without violation of any safety constraints even in cases where human external forces incidentally force the robot to violate constraints. The effectiveness of our approach is demonstrated in simulation studies on a two-link planar robot manipulator.

\end{abstract}

%%%%%%%%%%%%%%%%%%%%%%%%%%%%%%%%%%%%%%%%%%%%%%%%%%%%%%%%%%%%%%%%%%%%%%%%%%%%%%%%
\section{INTRODUCTION}\label{sec1}
%%%%%%%%%%%%%%%%%%%%%%%%%%%%%%%%%%%%%%%%
%  Motivation
%%%%%%%%%%%%%%%%%%%%%%%%%%%%%%%%%%%%%%%%
\subsection{Motivation} 
The past few decades have seen the rapid development in physical human-robot collaboration (pHRC) which is an increasingly important area in robotics. In the past, robots and human operators have been organised in separate areas to ensure safety \cite{RN532}.  
However, the physical collaboration and interaction between humans and robots are unavoidable in some specific scenarios, such as rehabilitation robots \cite{RN100} which guide a patient's arm and move with a natural fluidity with human movements.
In addition to this, the pHRC of industrial robots has been attracting a lot of interest in both academia and industry. 
Such a human-in-the-loop system makes full use of the reasoning capabilities of human workers and the high precision and endurance of robots \cite{RN546}, and therefore can finish much more complicated tasks compared to fully automated systems. 
To achieve a so-called \emph{compliance} in the physical interaction between human and robot, there is a vast body of work on impedance/admittance based compliance control in robotics, by which the movement of robots can be smoother, softer and more human-friendly.
However, bringing such collaborative robots into real-world use also requires safety guarantees since the robot and human share the same workspace. 
It is imperative that robots are capable of enforcing safety constraints with a human-like smooth behaviour when collaborating with a human partner. 
Safety should be strictly guaranteed even in some unexpected emergency situations such as human pushing the robot towards obstacles or beyond the workspace boundary by accident.

\begin{figure}[t]
	\centering
	\includegraphics[width=3.4in]{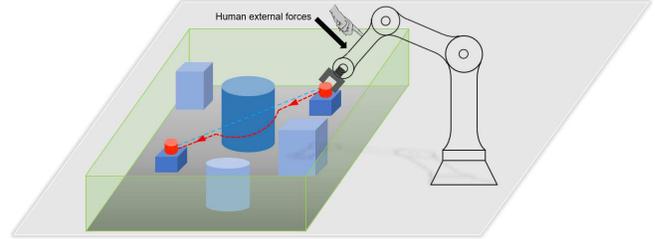}
	\caption{Motivating example scenario for a human robot force interaction with safety requirements.}
	\label{fig_1}
\end{figure}

\begin{figure}[t]
	\centering
	\includegraphics[width=3.4in]{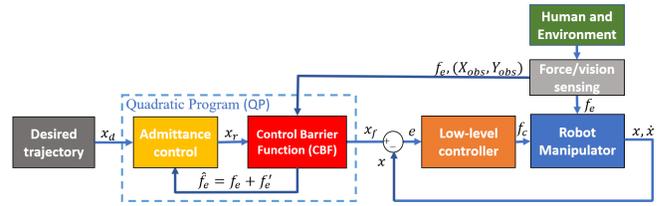}
	\caption{Structure of the proposed adaptive admittance control framework.}
	\label{fig_2}
\end{figure}

To that end, we seek to enforce the safety-critical constraints of admittance control in pHRC. In particular, we consider two position-based constraints, workspace boundary and obstacle avoidance, which are the most practical safety constraints in reality. As depicted in Fig. \ref{fig_1}, the trajectory of the robot's end-effector should always be constrained within the workspace boundary and keep a specific distance from obstacles. 
The proposed system implements a two-layered control framework: the high-level trajectory planning modifies the original desired trajectory in real-time. The ECBFs-QP based framework is applied to provide optimal virtual force feedback for admittance control as a compensator of the external human forces. Then, the low-level controller tracks such trajectories to achieve safety and human-friendly behaviour for robots, as depicted in Fig. \ref{fig_2}.
%%%%%%%%%%%%%%%%%%%%%%%%%%%%%%%%%%%%%%%%
%  Related Work
%%%%%%%%%%%%%%%%%%%%%%%%%%%%%%%%%%%%%%%%
\subsection{Related Work}
The most common control methods in robotics include position control \cite{RN88015}, force control \cite{844062}, hybrid position/force control \cite{uchiyama1987hybrid} and impedance/admittance based compliance control \cite{RN104}. Pure position control, whose task is to follow a specific trajectory as accurately as possible, rejects external human forces as disturbances. Therefore, it is not suitable in pHRC. 
In contrast, pure force control only tracks the given forces, and therefore cannot guarantee the position without contact with the external environment \cite{RN531}.
As a trade-off, impedance/admittance based compliance control relates both position and force. The contact point between the robot and human is assumed to be a mass-spring-damper (MSD) system whose dynamic behaviour acts like a human motion mechanism. 
Subsequently, many advanced control methods have emerged based on impedance/admittance control, such as adaptive impedance/admittance control \cite{kelly1989adaptive}\cite{colbaugh1993direct}, hybrid impedance/admittance control \cite{anderson1988hybrid}\cite{liu1991robust}, robust impedance/admittance control \cite{liu1991robust}\cite{chan1991robust} and learning impedance/admittance control \cite{RN1999}\cite{RN343}, etc. 
Note that the impedance and admittance are two opposite notions in the MSD system. 
The system is regarded as admittance when the input of the system is force and the output is position, while it is impedance when the input of the system is position and the output is force. 
In this work, we seek to get optimal interaction force feedback based on ECBFs-QP for compliance control to generate the trajectory which guarantees both safety and compliance. Therefore, we apply admittance control in our proposed framework.

In addition to compliance, another major factor of collaborative robots that hinders their use in real pHRC is safety. Previous solutions in safety-critical control in robotics mainly include: (i) path planning which aims to derive a collision-free trajectory to enforce safety, and (ii) introducing a safety filter which modifies the input of the controller to guarantee safety. Path planning mainly includes heuristic-based methods \cite{al2003efficient} \cite{li2005mobile} and potential field based methods \cite{1225434} \cite{RN449}. In the work of AI-Khatib and Saade \cite{al2003efficient}, a data-driven fuzzy approach is developed for a mobile robot to achieve path planning for moving obstacles. 
In short, heuristic-based methods have distinct advantages but the major drawbacks are response time and high computational complexity \cite{RN451}. 
In the work of Tang et al. \cite{1225434}, an artificial potential field (APF) based path planning was proposed for obstacle avoidance without the problem of local minimums. 
However, a work of Singletary et al. \cite{RN449} compared the performance of APF and control barrier functions (CBFs), and showed that CBFs outperformed existing APF-based algorithms. Therefore, CBFs have been an increasingly popular technique in the form of a safety filter. 
Using a quadratic program (QP), CBFs based safety filters can combine with an arbitrary nominal control law to enforce safety \cite{RN444} \cite{RN999}. 
Importantly, the CBFs-QP based framework can mediate the extent to which different constraints are met when these objectives are in conflict \cite{RN477}. 
Such a property makes CBFs ideal for pHRC since the safety of a robotics system should be a hard constraint because any violation is not acceptable, while tracking performance for the desired trajectory to finish tasks should be a soft constraint. CBFs are capable of finding a trade-off of each component to achieve an ideal behaviour. Another extension of CBFs called exponential control barrier functions (ECBFs) was developed in the work of Nguyen and Sreenath \cite{nguyen2016exponential} \cite{RN501}. which aims to handle the safety constraints with systems of higher relative degree.

So far, only a few papers have considered both compliance and safety in pHRC. Noteworthy among these is the work of Engelbrecht et al. \cite{RN588}, which proposes a novel adaptive virtual impedance algorithm for obstacle avoidance. Unlike traditional impedance control, the main difference in this work is that the impedance control is compensated by force feedback which is generated by APF. The APF provides both repulsive forces and attractive forces as adaptive feedback for impedance control to enable robots to navigate with constraints.
%%%%%%%%%%%%%%%%%%%%%%%%%%%%%%%%%%%%%%%%
%  Contribution
%%%%%%%%%%%%%%%%%%%%%%%%%%%%%%%%%%%%%%%%
\subsection{Contribution}
Motivated by the above discussion, in this letter, we apply the ECBFs-QP framework to provide virtual force feedback for admittance control. Such virtual force is an adaptive compensator for the external human forces. As can been seen in Fig. \ref{fig_2}, the optimal interaction force $\hat{f_e}$ is derived from the sum of external human forces $f_e$ and the compensated forces $f_{e}^{'}$. When the system tends to approach the unsafe set, $\hat{f_e}$ modifies the desired trajectory $x_r$ to $x_f$ for the low-level controller to track and bring the system back to the safe set. The major contribution of this work can be highlighted in a comparison with other approaches as follows:
\begin{enumerate}
	\item Compared with traditional path planning \cite{al2003efficient} \cite{li2005mobile}, the proposed approach integrates an adaptive force compensator with admittance control to modify the desired trajectory. Therefore, the proposed control framework has higher robustness to external human forces, as well as shorter response time and lower computational complexity.
	\item Compared with existing safety filters \cite{RN444} \cite{RN999}, the proposed approach seeks to enforce safety in the \emph{force} domain. The main benefit is the improvement in the robot's human-friendly motion behaviour in the physical human robot force interaction, while simultaneously guaranteeing the safety constraints.
\end{enumerate}

%%%%%%%%%%%%%%%%%%%%%%%%%%%%%%%%%%%%%%%%
%  Organization
%%%%%%%%%%%%%%%%%%%%%%%%%%%%%%%%%%%%%%%%
\subsection{Organization}
The remainder of the letter is organised as follows. Section \ref{sec2} provides the necessary background on the robot manipulator model and ECBFs. Section \ref{sec3} presents the formulation of the proposed control framework subject to two position-based constraints: workspace constraints and obstacle avoidance. In Section \ref{sec4}, the formulation of a low-level controller is given based on our previous works. The simulation results are demonstrated on a two-link planar robotic manipulator in Section \ref{sec5}. Finally, Section \ref{sec6} discusses the conclusions and directions for future works.     
%%%%%%%%%%%%%%%%%%%%%%%%%%%%%%%%%%%%%%%%%%%%%%%%%%%%%%%%%%%%%%%%%%%%%%%%%%%%%%%%%%%%%%%%%%%%%%%%%%%%%%%%%%%%%%
%  Preliminaries
%%%%%%%%%%%%%%%%%%%%%%%%%%%%%%%%%%%%%%%%%%%%%%%%%%%%%%%%%%%%%%%%%%%%%%%%%%%%%%%%%%%%%%%%%%%%%%%%%%%%%%%%%%%%%%
\section{Preliminaries}\label{sec2}
%%%%%%%%%%%%%%%%%%%%%%%%%%%%%%%%%%%%%%%%
%  Model
%%%%%%%%%%%%%%%%%%%%%%%%%%%%%%%%%%%%%%%%
\subsection{Model}
In this section, we begin by briefly introducing the model that captures the dynamics of the robot manipulator. The joint space dynamics of a robot manipulator can be written as:
\begin{equation}
	\label{eq1}
	M\left( q \right) \ddot{q}+C\left( q,\dot{q} \right) \dot{q}+G\left( q \right) +F\left( \dot{q} \right) =\tau _c +\tau _e
\end{equation}
where $q$, $\dot{q}$, $\ddot{q}$ are the joint positions, velocities and accelerations in joint space. $M\left( q \right)$, $C\left( q,\dot{q} \right)$, $G\left( q \right)$ and $F\left( \dot{q} \right)$ are mass, Coriolis force and centrifugal force, gravity and friction coefficient matrices, respectively. $\tau_c$ denotes the control torque, and $\tau_e$ is the external torque from the environment. In the design of a robotic control system, the system needs to be formulated in Cartesian space when evaluating the trajectory tracking performance. The time-varying transformation between the joint velocities and Cartesian velocities of the robot manipulator can be written as:
\begin{equation}
	\label{eq2}
	\dot{x}=J\left( q \right) \dot{q}
\end{equation}
where $J\left( q \right)$ is the Jacobian of the robot manipulator. By using (\ref{eq2}), we can then transfer the joint space dynamics (\ref{eq1}) into Cartesian space as:
\begin{equation}
	\label{eq3}
	M_x\ddot{x}+C_x\dot{x}+G_x+F_x=f_c+f_e
\end{equation}
where $x$, $\dot{x}$, $\ddot{x}$ is the position, velocity and acceleration of the robot joints in Cartesian space. $f_c=J^{-T}\left( q \right) \tau _c$, and $f_e=J^{-T}\left( q \right) \tau _e$ present the control forces and the external human forces, respectively. The coefficient matrices are given by $M_x=J^{-T}\left( q \right) M\left( q \right) J^{-1}\left( q \right) $, $C_x=J ^{-T}\left( q \right)\left( C\left( q,\dot{q} \right) -M\left( q \right) J^{-1}\left( q \right) \dot{J}\left( q \right) \right) J^{-1}\left( q \right) $, $G_x=J^{-T}\left( q \right) G\left( q \right)$ and $\ F_x=J^{-T}\left( q \right) F\left( \dot{q} \right) $. Note that another benefit of transferring joint space dynamic into Cartesian space is that we can relate the external force $f_e$ (rather than torque $\tau_e$) in (\ref{eq3}) directly to the forces in the equation of admittance control, which will be mentioned in later sections.
%%%%%%%%%%%%%%%%%%%%%%%%%%%%%%%%%%%%%%%%
%  ECBF
%%%%%%%%%%%%%%%%%%%%%%%%%%%%%%%%%%%%%%%%
\subsection{Exponential Control Barrier Function}

In this section, we revisit the mathematical background of exponential control barrier functions. Consider a nonlinear control system of the form:
\begin{equation}
	\left\{ \begin{array}{l}
		\dot{x}=f\left( x,u \right) \\
		y=\zeta \left( x \right)\\
	\end{array} \right. 
	\label{e1}
\end{equation}
where $x\in \mathbb{R}^n$ is the system state, $u\in U\subset \mathbb{R}^m$ is the admissible control input, and $y\in \mathbb{R}^m$ is the control output. $f:\mathbb{R}^m\rightarrow \mathbb{R}^n$ is locally Lipschitz. The primary focus of control barrier functions is to force system states to remain in a given safe set $C$ which is defined as the superlevel set of a continuously differentiable function  $h:\mathbb{R}^n\rightarrow \mathbb{R}$:
\begin{equation}
	C=\left\{ x\in \mathbb{R}^n:h\left( x \right) \ge 0 \right\} 
	\label{e5}
\end{equation}
\begin{equation}
	\partial C=\left\{ x\in \mathbb{R}^n:h\left( x \right) =0 \right\} 
	\label{e6}
\end{equation}
\begin{equation}
	\mathrm{Int}\left( C \right)  =\left\{ x\in \mathbb{R}^n:h\left( x \right) >0 \right\} 
	\label{e7}
\end{equation}
where $h\left(x\right)$ is called the constraint function. $h\left(x\right) >0$ indicates safety, while $h\left(x\right) <0$ indicates violation of the safety constraints.  $C$ denotes the safe set, $\partial C$ denotes the boundary of the set $C$, and $\mathrm{Int}\left( C \right) $ denotes the interior of the set $C$. The mathematical tools that underpin application of control safety features such as CBFs are based on results from the well-known Nagumo's Theorem \cite{RN235467} which provides the necessary sufficient conditions of the invariant set:
\begin{equation}
	\dot{h}\left( x \right) \ge 0, \forall x\in \partial C
\end{equation}

Using these conditions, enforcement of the safety of a control system can be converted to another question: \emph{what sufficient conditions that need to be imposed on $h\left(x\right)$ so that $\mathrm{Int}\left( C \right) $ is forward invariant?} CBFs provide a solution to this question and therefore can enforce the \emph{safety} of a control system. 

\noindent\textbf{\emph{Definition 1}}. Let $u \in U$ be a control value for the system (\ref{e1}). For any initial states $x_0:=x\left(t_0\right)$, $x\left(t\right)$ is the unique solution to (\ref{e1}) in the maximum time interval $T\left( x_0 \right)$. The set $S$ is \emph{forward invariant} with respect to the control value $u$ if for every $x_0\in S$, $x\left(t \right)\in S$ for all $t\in T\left(x_0\right)$. The control system (\ref{e1}) is \emph{safe} with respect to the set $S$ if the set $S$ is forward invariant.

Essentially, there are two types of CBFs: one is \emph{reciprocal control barrier functions} which tend to infinity on the set boundary, and one is \emph{zeroing control barrier functions} which vanish on the set boundary. 

1) \emph{Reciprocal control barrier functions (RCBFs)} \cite{RN444}: A RCBF $B\left(x\right)$ which tends to infinity on the set boundary $\partial C$, should satisfy the following important properties:
\begin{equation}
	\inf_{x\in \mathrm{Int}\left( C \right) }B\left( x \right) \ge 0,\ \ \lim_{x\rightarrow \partial C}B\left( x \right) =\infty 
	\label{p1}
\end{equation}

\noindent\textbf{\emph{Definition 2}}. Consider a control system in the form of (\ref{e1}). For a given safe set $C$, admissible control input set $U$ and a continuously differentiable function $h\left(x\right)$,  $B\left(x\right)$ is a \emph{reciprocal control barrier function} if there exist class $\mathcal{K}$ functions $\alpha_1$, $\alpha_2$, $\alpha_3$ such that for all $x\in \mathrm{Int}\left( C \right) $, $B\left(x\right)$ satisfies following conditions:
\begin{equation}
	\frac{1}{\alpha _1\left( h\left( x \right) \right)}\le B\left( x \right) \le \frac{1}{\alpha _2\left( h\left( x \right) \right)}
\end{equation}
\begin{equation}
	\inf_{u\in U}\left[ \mathscr{L}_fB\left( x ,u\right) -\alpha _3\left( h\left( x \right) \right) \right] \le 0
\end{equation}
where $\mathscr{L}_fB\left( x \right)$ is the \emph{Lie derivative} of $B\left( x \right)$ with respect to the vector field $f$. To satisfy the properties in (\ref{p1}), reciprocal control barrier function candidates are usually selected as the form of the inverse-type barrier candidate $B\left( x \right) =\frac{1}{h\left( x \right)}$, logarithmic barrier function candidate $B\left( x \right) =-\log \left( \frac{h\left( x \right)}{1+h\left( x \right)} \right)$ and so on.

\noindent\textbf{\emph{Lemma 1}} \cite{RN444}. Consider a control system in the form of (\ref{e1}). For a given safe set $C$ defined by (\ref{e5})-(\ref{e7}), admissible control input set $U$ and a RCBF $B\left(x\right)$, any locally Lipschitz continuous control input $u\left(x\right)\in U$ such that $u\left(x\right) \in K_{rcbf}\left( x \right) =\left\{ u\in U:\mathscr{L}_fB_r\left( x,u \right)-\alpha _3\left( h\left( x \right) \right) \le 0 \right\} 
$ will render the set $\mathrm{Int}\left( C \right) $ forward invariant.

\noindent\textbf{\emph{Definition 3}}. Consider a continuous function $\alpha :\left[ 0,\infty  \right) \rightarrow \left[ 0,\infty  \right)$. It is said the $\alpha$ is a \emph{class $\mathcal{K}$ function} if it is strictly increasing and $\alpha\left(0\right)=0$. Based on class $\mathcal{K}$ functions, \emph{extended class $\mathcal{K}$ functions} are defined on the entire real line $\mathbb{R}=\left( -\infty ,\infty \right) $.

\noindent\textbf{\emph{Definition 4}}. Consider a control system in the form of (\ref{e1}). For a scalar function $B\left(x\right)$, the directional derivatives of $B\left(x\right)$ with respect to the vector field $f\left(x,u\right)$ are called the \emph{Lie derivatives} of $B\left(x\right)$ along $f\left(x,u\right)$ and are denoted by:
\begin{equation}
	\mathscr{L}_fB\left( x ,u\right) =\frac{\partial B\left( x \right)}{\partial x}f\left( x,u \right)
\end{equation}

2) \emph{Zeroing control barrier functions (ZCBFs)} \cite{RN444}: Since the unbounded values of $B\left(x\right)$ may be undesirable in practical implementations, employing ZCBFs which vanish on the set boundary $\partial C$ give a solution to this problem. A ZCBF $h\left(x\right)$ satisfies the following important properties:
\begin{equation}
	\inf_{x\in \mathrm{Int}\left( C \right) }h\left( x \right) \ge 0,\ \ \lim_{x\rightarrow \partial C}h\left( x \right) =0
\end{equation}
\noindent\textbf{\emph{Definition 5}}. Consider a control system in the form of (\ref{e1}). For a given safe set $C$, admissible control input set $U$ and a continuously differentiable function $h\left(x\right)$,  $h\left(x\right)$ is a \emph{zeroing control barrier functions} if there exist an extended class $\mathcal{K}$ function $\alpha$ such that for all $x\in \mathrm{Int}\left( C \right) $, $h\left(x\right)$ satisfies the following conditions:
\begin{equation}
	\sup_{u\in U}\left[ \mathscr{L}_fh\left( x,u \right)-\alpha \left( h\left( x \right) \right) \right] \ge 0
	\label{s1}
\end{equation}
\noindent\textbf{\emph{Lemma 2}} \cite{RN444}: Consider a control system in the form of (\ref{e1}). For a given safe set $C$ defined by (\ref{e5})-(\ref{e7}), admissible control input set $U$ and a ZCBF $h\left(x\right)$, any locally Lipschitz continuous control input $u\left(x\right)\in U$ such that $u\left(x\right) \in K_{zcbf}\left( x \right) =\left\{ u\in U:\mathscr{L}_fh\left( x,u \right)-\alpha \left( h\left( x \right) \right) \ge 0 \right\} $ will render the set $\mathrm{Int}\left( C \right) $ forward invariant.

\noindent\textbf{\emph{Remark 1}}. Note that usually the $\alpha$ in (\ref{s1}) is the extended class $\mathcal{K}$ function. There is a special case in which the term $\alpha\left(h\left(x\right)\right)$ is replaced by $\lambda h\left(x\right)$, where $\lambda$ is a positive real number which is selected based on experience. In this letter, we apply this special case of ZCBFs in our proposed control framework.

The first order Lie derivatives of CBFs with higher relative degree than one do not depend explicitly on the control input offering no way to calculate what control changes need to be applied to enforce the safety condition. Therefore, an extension of the CBFs, called \emph{exponential control barrier function}, is proposed for dealing with higher relative degree constraint functions \cite{nguyen2016exponential} \cite{RN501}. In this letter, we consider two position-based constraints with relative degree two, therefore, we apply ECBFs in our proposed control framework. 

\noindent\textbf{\emph{Definition 6}}. For a given control system (\ref{e1}), safe set $C$ defined by (\ref{e5})-(\ref{e7}), admissible control input set $U$ and a ZCBF $h\left(x\right)$ which has relative degree $r$, then $h\left(x\right)$ is an \emph{exponential control barrier function} if there exists a $K\in \mathbb{R}^{r}$ such that for any $x\in C$, the following condition is satisfied:
\begin{equation}
	\inf_{u\in U}\left[ \mathscr{L}_{f}^{r}h\left( x,u \right) +K\xi \left( x \right) \right] \ge 0
	\label{key}
\end{equation}
where $\xi \left( x \right) =\left[ h\left( x \right) ,\mathscr{L}_fh\left( x \right) ,\mathscr{L}_{f}^{2}h\left( x \right) ,...,\mathscr{L}_{f}^{r-1}h\left( x \right) \right] ^T$ and $K=\left[k^1, k^2,...k^{r-1}\right]$. Equation (\ref{key}) provides the sufficient condition for safety in our proposed framework.

\noindent\textbf{\emph{Lemma 3}} \cite{RN501}. Consider a control system in the form of (\ref{e1}). For a given safe set $C$ defined by (\ref{e5})-(\ref{e7}), admissible control input set $U$ and a ECBF $h\left(x\right)$ with relative degree $r$, any locally Lipschitz continuous control input $u\left(x\right)\in U$ such that $u\left(x\right) \in K_{ecbf}\left( x \right) =\left\{ u\in U:\mathscr{L}_{f}^{r}h\left( x,u \right) +K\xi \left( x \right) \ge 0 \right\} $ will render the set $\mathrm{Int}\left( C \right) $ forward invariant.            
%%%%%%%%%%%%%%%%%%%%%%%%%%%%%%%%%%%%%%%%%%%%%%%%%%%%%%%%%%%%%%%%%%%%%%%%%%%%%%%%%%%%%%%%%%%%%%%%%%%%%%%%%%%%%%
%  Adaptive admittance control for safety-critical control
%%%%%%%%%%%%%%%%%%%%%%%%%%%%%%%%%%%%%%%%%%%%%%%%%%%%%%%%%%%%%%%%%%%%%%%%%%%%%%%%%%%%%%%%%%%%%%%%%%%%%%%%%%%%%%
\section{Adaptive admittance control for safety-critical control}\label{sec3}
%%%%%%%%%%%%%%%%%%%%%%%%%%%%%%%%%%%%%%%%
%  Admittance control reconstruction
%%%%%%%%%%%%%%%%%%%%%%%%%%%%%%%%%%%%%%%%
\subsection{Admittance control reconstruction}
\begin{figure}[h]
	\centering
	\includegraphics[width=1.7in]{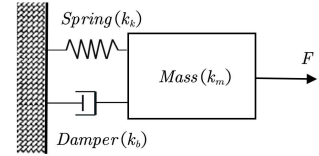}
	\caption{ The mass-spring-damper system.}
	\label{p2}
\end{figure}

To provide compliance for physical human-robot force interaction, the contact point between the human and the robot is modelled as a mass-spring-damper system to imitate human muscle mechanisms, as depicted in Fig. \ref{p2}. The virtual mass, spring, and damper ensure that the interaction forces are elastic and never vibrate at the contact point. The dynamics for a robot manipulator rendering an impedance can be written as:
\begin{small}
\begin{equation}
	\label{eq10}
	k_{m_i}\left( \ddot{x}_{r_i}-\ddot{x}_{d_i} \right) +k_{b_i}\left( \dot{x}_{r_i}-\dot{x}_{d_i} \right) +k_{k_i}\left( x_{r_i}-x_{d_i} \right) =f_{e_i}
\end{equation}
\end{small}
where $k_{m_i}$, $k_{b_i}$ and $k_{k_i}$ are the mass, spring and damping coefficients, respectively. $i$ is the number of degree of freedom (DOF) of the robot manipulator. $x_{d_i}$ is the desired trajectory that has been determined in advance to finish the task. $x_{r_i}$ is the reference trajectory modified by the effect of the external force by using admittance control. $f_{e_i}$ is the external human force.

Let $x_{1_i}=x_{r_i}$, $x_{2_i}=\dot{x}_{r_i}$, $u_i=f_{e_i}$, from (\ref{eq10}), we have:
\begin{small}
\begin{equation}
	\left\{ \begin{array}{l}
		\dot{x}_{1_i}=x_{2_i}\\
		\dot{x}_{2_i}=f_i\left( x_i \right) +g_i\left( x_i \right) u_i\\
	\end{array} \right. 
	\label{s2}
\end{equation}
\end{small}
where  $f_i\left( x_i \right)$ and $g_i\left( x_i \right)$ are written as $f_i\left( x_i \right) =-\frac{1}{k_{m_{\text{i}}}}\left[ k_{b_i}\left( x_{2_i}-\dot{x}_{d_i} \right) +k_{k_i}\left( x_{1_i}-x_{d_i} \right) -k_{m_i}\ddot{x}_{d_i} \right]$ and 
$g_i\left( x_i \right) =\frac{1}{k_{m_i}}$.

%%%%%%%%%%%%%%%%%%%%%%%%%%%%%%%%%%%%%%%%
%  Workspace constraints
%%%%%%%%%%%%%%%%%%%%%%%%%%%%%%%%%%%%%%%%
\subsection{Workspace constraints}
In this section, we present the application of ECBFs for admittance system (\ref{s2}) with workspace constraints. We assume the end-effector of the robot is required to be constrained within a virtual box that defines its desired working area. For this case, there will be an upper boundary and a lower boundary for each sate in (\ref{s2}). These are:
\begin{equation}
	x_{\min _i}\le x_{r_i}\le x_{\max _i}
\end{equation}

Then, the constraint function $h\left(x\right)$ for the upper and lower boundary of each state can be designed as:
\begin{small}
	\begin{equation}
		\begin{split}
			h_{\max _i}\left( x \right) =\left( x_{1_i}-x_{\max _i} \right) ^2-r^2 \\
			h_{\min _i}\left( x \right) =\left( x_{\min _i}-x_{1_i} \right) ^2-r^2
		\end{split}
	\label{h1h2}
	\end{equation}
\end{small}
where $r$ denotes the safe distance between the end-effector and workspace boundary. The safe set is defined as $	C=\left\{ x\in \mathbb{R}^n:h_{\max _i}\ge 0\ \&\ h_{\min _i}\ge 0 \right\}$. The Lie derivative of the constraint function (\ref{h1h2}) with respect to $f\left(x,u\right)$ is:
\begin{small}
	\begin{equation*}
	\begin{split}
		\mathscr{L}_fh_{\max _i}\left( x \right) =2\left( x_{1_i}-x_{\max _i} \right) \dot{x}_{1_i}=2\left( x_{1_i}-x_{\max _i} \right) x_{2_i}\ \  \\
		\mathscr{L}_fh_{\min _i}\left( x \right) =-2\left( x_{\min _i}-x_{1_i} \right) \dot{x}_{1_i}=-2\left( x_{\min _i}-x_{1_i} \right) x_{2_i}
	\end{split}
	\end{equation*}
\end{small}

Since (\ref{h1h2}) have relative degree two, the second-order Lie derivative of the constraint functions are:
\begin{small}
\begin{equation}
	\begin{aligned}
	\mathscr{L}_{f}^{2}h_{\max _{\text{i}}}\left( x,u \right) &=2\left( x_{1_i}-x_{\max _i} \right) \dot{x}_{2_i}+2x_{2_i}\\
	&=2\left( x_{1_i}-x_{\max _i} \right) \left( f_i\left( x \right) +g_i\left( x \right) u_i \right) +2x_{2_i}\\
	\mathscr{L}_{f}^{2}h_{\min _i}\left( x,u \right) &=-2\left( x_{\min _i}-x_{1_i} \right) \dot{x}_{2_i}+2x_{2_i}\\
	&=-2\left( x_{\min _i}-x_{1_i} \right) \left( f_i\left( x \right) +g_i\left( x \right) u_i \right) +2x_{2_i}
	\end{aligned}
	\label{c7c8}
\end{equation}
\end{small}

According to (\ref{key}), to achieve the forward invariance of the safe set $C$, the following inequalities should be satisfied:
\begin{small}
\begin{equation}
	\begin{split}
	\mathscr{L}_{f}^{2}h_{\max _i}\left( x,u \right) +K_{\max_i}\xi _{\max_i}\left( x \right) \ge 0\\
	\mathscr{L}_{f}^{2}h_{\min _i}\left( x,u \right) +K_{\min_i}\xi _{\min_i}\left( x \right) \ge 0
	\label{c1c2}
	\end{split}
\end{equation}
\end{small}
where $\xi _{\max _i}\left( x \right) =\left[ h_{\max _i}\left( x \right) ,\mathscr{L}_fh_{\max _i}\left( x \right) \right] ^T$ and $\xi _{\min _i}\left( x \right) =\left[ h_{\min _i}\left( x \right) ,\mathscr{L}_fh_{\min _i}\left( x \right) \right] ^T$. The parameter $K_{\max_i}=\left[k_{\max_1},k_{\max_2}\right]$ and $K_{\min_i}=\left[k_{\min_1},k_{\min_2}\right]$ denote the positive coefficients of the ECBFs. The selection of these parameters depends on the dynamics of the robot manipulator. These parameters are selected based on a trial-and-error procedure and by experience. Then, the Quadratic Program is applied here to derive minimal modifying forces that maintain the safety conditions subject to the conditions (\ref{c1c2}):
\begin{small}
\begin{equation}
	\begin{aligned}
	\hat{f}_{e}^{'}&=\underset{\hat{f}_{e_i}}{arg\min}\lVert \hat{f}_e-f_e \rVert ^2\\
	s.t.\ \ \ \mathscr{L}_{f}^{2}h_{\max _i}&\left( x,u \right) +K_{\max_i}\xi _{\max_i}\left( x \right) \ge 0\\
	\mathscr{L}_{f}^{2}h_{\min _i}&\left( x,u \right) +K_{\min_i}\xi _{\min_i}\left( x \right) \ge 0
	\label{key1}
	\end{aligned}
\end{equation}
\end{small}
where $\hat{f}_{e}$ is the minimal modifying force we wish to find, $f_e$ is the external human force and $\hat{f}_{e}^{'}$ is the compensated force. By using (\ref{key1}), the minimum $\hat{f}_{e}^{'}$ is picked up based on the conditions (\ref{c1c2}). The final output from the QP $\hat{f}_{e}^{'}$ compensates the human forces as an add-on to $f_e$, resulting in the optimal interaction force $\hat{f}_{e}$ which enforces the safety of system (\ref{s2}). In addition, the natural of system (\ref{s2}) also guarantees the compliance of $\hat{f}_{e}$. We define the constraints within the QP in the form:
\begin{equation}
	Au-b\le  0
	\label{c5}
\end{equation}

Let $u_{nom}=f_e$, $u_{safe}=\hat{f}_{e}$. Taking maximum boundary as an example, writing (\ref{c1c2}) in the form of (\ref{c5}), we have:
\begin{small}
	\begin{equation*}
		\begin{split}
		-\mathscr{L}_{f}^{2}h_{\max _i}\left( x,\left( u_{nom}-u_{safe} \right) \right) \le K_{\max _i}\xi _{\max _i}\left( x \right)\ \ \ \ \ \ \  \\
		\mathscr{L}_{f}^{2}h_{\max _i}\left( x,u_{safe} \right) \le \mathscr{L}_{f}^{2}h_{\max _i}\left( x,u_{nom} \right) +K_{\max _i}\xi _{\max _i}\left( x \right) 
		\end{split}
	\end{equation*}
\end{small}

According to (\ref{c7c8}), we have:
\begin{small}
\begin{equation*}
	\begin{split}
	2\left( x_{1_i}-x_{\max _i} \right) g_i\left( x \right)u_{safe_i} \le \ \ \ \ \ \ \ \ \ \ \\
	2\left( x_{1_i}-x_{\max _i} \right) g_i\left( x \right) u_{nom_i}+K_{\max _i}\xi _{\max _i}\left( x \right) 	
	\end{split}
\end{equation*}
\end{small}

Similarly, for minimum boundary we have:
\begin{small}
\begin{equation*}
	\begin{split}
		-2\left( x_{1_i}-x_{\min _i} \right) g_i\left( x \right)u_{safe_i} \le \ \ \ \ \ \ \ \ \ \ \\
		-2\left( x_{1_i}-x_{\min _i} \right) g_i\left( x \right) u_{nom_i}+K_{\min _i}\xi _{\min _i}\left( x \right) 	
	\end{split}
\end{equation*}
\end{small}

Therefore, when applying (\ref{key1}) for practical application, $A_{w_i}$ and $b_{w_i}$ for the i-th DOF of the robot manipulator can be written as:
\begin{small}
\begin{equation*}
	\begin{split}
	A_{w_i}=\left[ \begin{matrix}
		2\left( x_{1_i}-x_{\max _i} \right) g_i\left( x \right) &		0\\
		0 &	-2\left( x_{\min _i}-x_{1_i} \right) g_i\left( x \right)\\
	\end{matrix} \right] \\
	b_{w_i}=\left[ \begin{array}{c}
		2\left( x_{1_i}-x_{\max _i} \right) g_i\left( x \right) u_{nom_i}+K_{\max _i}\xi _{\max _i}\left( x \right)\\
		-2\left( x_{1_i}-x_{\min _i} \right) g_i\left( x \right) u_{nom_i}+K_{\min _i}\xi _{\min _i}\left( x \right)\\
	\end{array} \right] 
	\end{split}
\end{equation*}
\end{small}

%%%%%%%%%%%%%%%%%%%%%%%%%%%%%%%%%%%%%%%%
%  Obstacle avoidance
%%%%%%%%%%%%%%%%%%%%%%%%%%%%%%%%%%%%%%%%
\subsection{Obstacle avoidance}
In this section, we present the application of ECBFs for admittance system (\ref{s2}) with obstacle constraints. We assume the end-effector of the robot should always keep a safe distance from the obstacles. For this case, the constraint function is designed as:
\begin{small}
\begin{equation}
	h_{obs}\left( x \right) =\sum_1^i{\left( x_{1_i}-x_{obs_i} \right) ^2-r^2}
	\label{h3}
\end{equation}
\end{small}
where $x_{obs_i}$ denotes the position of obstacle in the Cartesian space and $r$ denotes the safe distance. The safe set is define as $C=\left\{ x\in \mathbb{R}^n:h_{obs}\left( x \right)\ge 0\right\}$. The first and second order Lie derivative of the constraint function (\ref{h3}) with respect to the $f\left(x,u\right)$ are:
\begin{small}
	\begin{equation}
		\begin{split}
			\mathscr{L}_fh_{obs}\left( x \right) =\sum_1^i{\left( 2\left( x_{1_i}-x_{obs_i} \right) \dot{x}_{1_i} \right) =}\sum_1^i{\left( 2\left( x_{1_i}-x_{obs_i} \right) x_{2_i} \right)}\\
			\mathscr{L}_{f}^{2}h_{obs}\left( x,u \right) =\sum_1^i{\left( 2\left( x_{1_i}-x_{obs_i} \right)
				 \left( f_i\left( x \right) +g_i\left( x \right) u_i \right) +2x_{2_i} \right)}
			\label{c9}
		\end{split}
	\end{equation}
\end{small}

According to (\ref{key}), to achieve the forward invariance of the safe set $C$, the following inequalities should be satisfied:
\begin{small}
\begin{equation}
	\mathscr{L}_{f}^{2}h_{obs}\left( x,u \right) +K_{obs}\xi_{obs} \left( x \right) \ge 0
	\label{c6}
\end{equation}
\end{small}
where $K_{obs}=\left[k_1,k_2\right]$, $\xi_{obs} \left( x \right) =\left[ h\left( x \right) ,\mathscr{L}_fh\left( x \right) \right] ^T$. Then, the Quadratic Program is applied to derive optimal interaction forces subject to the conditions (\ref{c6}):
\begin{small}
\begin{equation}
	\begin{aligned}
		\hat{f}_{e}^{'}=\underset{\hat{f}_{e_i}}{arg\min}\lVert \hat{f}_e-f_e \rVert ^2\ \ \ \ \ \ \ \\
		s.t.  \mathscr{L}_{f}^{2}h_{obs}\left( x,u \right) +K_{obs}\xi_{obs} \left( x \right) \ge 0
		\label{key2key7}
	\end{aligned}
\end{equation}
\end{small}

Similarly to the workspace constraints, when applying (\ref{key2key7}) for practical application, (\ref{key2key7}) can be rewritten as:
\begin{small}
\begin{equation*}
	\mathscr{L} _{f}^{2}h_{obs}\left( x,u_{safe} \right) \le \mathscr{L} _{f}^{2}h_{obs}\left( x,u_{nom} \right) +K_{obs}\xi_{obs} \left( x \right) 
\end{equation*}
\end{small}

According to (\ref{c9}), we have:
\begin{small}
\begin{equation*}
	\begin{aligned}
	\sum_1^i{2\left( x_{1_i}-x_{obs_i} \right) g_i\left( x \right) u_{safe_i}}\le \ \ \ \ \ \ \ \ \ \ \\
	\sum_1^i{2\left( x_{1_i}-x_{obs_i} \right) g_i\left( x \right) u_{nom_i}}+K_{obs}\xi_{obs} \left( x \right) 
\end{aligned}
\end{equation*}
\end{small}

Therefore,  when applying (\ref{key2key7}) as before within our QP constraints, $A$ and $b$ can be written as:
\begin{small}
\begin{equation*}
	\begin{split}
	A_{obs}=\left[ 2\left( x_{1_1}-x_{obs_1} \right) g_1\left( x \right) ,...,2\left( x_{1_i}-x_{obs_i} \right) g_i\left( x \right) \right] \\
	b_{obs}=\sum_1^i{2\left( x_{1_i}-x_{obs_i} \right) g_i\left( x \right) u_{nom_i}}+K_{obs}\xi_{obs} \left( x \right) 
	\end{split}
\end{equation*}
\end{small}
%%%%%%%%%%%%%%%%%%%%%%%%%%%%%%%%%%%%%%%%
%  Two constraints
%%%%%%%%%%%%%%%%%%%%%%%%%%%%%%%%%%%%%%%%
\subsection{Workspace and obstacle constraints simultaneously}
To achieve both workspace constraints and obstacle avoidance simultaneously, we then synthesize these two kinds of constraints into QP, we have:

\begin{equation}
	\begin{aligned}
		\hat{f}_{e}^{'}=\underset{\hat{f}_{e_i}}{arg\min}\lVert \hat{f}_e-f_e \rVert ^2\ \ \ \ \ \ \ \\
		s.t.\ \ \ \ \ \mathscr{L}_{f}^{2}h_{obs}\left( x,u \right) +K_{obs}\xi_{obs} \left( x \right) \ge 0 \\
		\mathscr{L}_{f}^{2}h_{\min _i}\left( x,u \right) +K_{\min_i}\xi _{\min_i}\left( x \right) \ge 0\\
		\mathscr{L}_{f}^{2}h_{\max _i}\left( x,u \right) +K_{\max_i}\xi _{\max_i}\left( x \right) \ge 0
		\label{key3456}
	\end{aligned}
\end{equation}

According to (\ref{c5}), the synthesized $A$ and $b$ of (\ref{key3456}) can be written as:
\begin{equation*}
	A=\left[ A_{obs}, A_{w_1}, A_{w_2},...,A_{w_i} \right]^T ,b=\left[ b_{obs},b_{w_1},b_{w_2},...,b_{w_i}\right]^T
\end{equation*}            

%%%%%%%%%%%%%%%%%%%%%%%%%%%%%%%%%%%%%%%%%%%%%%%%%%%%%%%%%%%%%%%%%%%%%%%%%%%%%%%%%%%%%%%%%%%%%%%%%%%%%%%%%%%%%%
% Low-Level control
%%%%%%%%%%%%%%%%%%%%%%%%%%%%%%%%%%%%%%%%%%%%%%%%%%%%%%%%%%%%%%%%%%%%%%%%%%%%%%%%%%%%%%%%%%%%%%%%%%%%%%%%%%%%%%
\section{Low-Level controller}\label{sec4}
As shown in Fig. 2, after the admittance control generated trajectory $x_r$ is adapted to satisfy the system safety constraints as trajectory $x_f$, the low-level controller will be designed to track the trajectory $x_f$. To guarantee the safety of the system, the low-level controller needs to guarantee that the system states will track trajectory $x_f$ precisely with very small or near zero offset. In this letter, we apply the fixed-time integral sliding mode controller (FxTISMC) \cite{RN488} as the low-level controller as it provides very high precision and tracking error convergence within a fixed time interval. The FxTISMC controller is designed as $u_c=u_0+u_{s}$, where $u_0$ is the nominal controller used to control the nominal component and $u_{s}$ is the compensating controller used to compensate for the model uncertainty and stabilise the system. 

1) \emph{Nominal controller}:

Letting $\eta _1=x$ and $\eta _2=\dot{x}$, according to the dynamics of the robot manipulator in Cartesian space (\ref{eq3}), the dynamics without uncertainties can be written as:
\begin{small}
\begin{equation*}
	\left\{ \begin{array}{l}
		\dot{\eta}_1=\eta _2\\
		\dot{\eta}_2=\varXi u+\varGamma \left( x, \dot{x} \right)\\
	\end{array} \right.  
\end{equation*}
\end{small}
where $\varGamma \left( x,\dot{x} \right) =M_x^{-1}\left( -C_x-G_x \right)$. Letting $s_1=\eta _1-x_r$, then, $\dot{s}_1=\eta_2-\dot{x_r}$, and the stabilizing function can designed as:
\begin{small}
\begin{equation*}
	\alpha _s=-\left( \lambda _1s_1+\lambda _2s_{1}^{\alpha}+\lambda _3s_{1}^{\beta} \right) +\dot{x}_r
\end{equation*}
\end{small}
where $\lambda _1$, $\lambda _2$, $\lambda _3$, $\alpha$, $\beta$ are all positive constants satisfying $0<\alpha<1$ and $\beta>1$. Letting $s_2=\eta _2-\alpha _s $, we have:
\begin{small}
\begin{equation*}
	\dot{s}_2=\dot{\eta}_2-\dot{\alpha}_s=\varXi u_0-\varGamma \left( x,\dot{x} \right) -\dot{\alpha}_s
\end{equation*}
\end{small}

The nominal controller is then designed as:
\begin{small}
\begin{equation*}
	u_0=\varXi ^{-1}\left( -\varGamma \left( x,\dot{x} \right) +\dot{\alpha}_s-\lambda _1s_2-\lambda _2s_{2}^{\alpha}-\lambda _3s_{2}^{\beta} \right) 
\end{equation*}
\end{small}

2) \emph{Compensating controller}:

Let the tracking error $e=x-x_r$, select the sliding variable as:
\begin{small}
\begin{equation*}
	s = e+\frac{1}{\kappa_1^{m}}\left[ \dot{e}+\kappa_2\left[ e \right] ^n \right] ^{\frac{1}{m}}
\end{equation*}
\end{small}
where $\kappa_ 1$, $\kappa_2$, $\kappa_ 3$, $\kappa_4$, $m$, $n$ are all constants satisfying $\kappa_ 1, \kappa_2, \kappa_3, \kappa_4>0$ , $0<m<1$ and $n>1$. The integral sliding surface is then selected as:
\begin{small}
\begin{equation*}
	\begin{split}
		\sigma \left( t \right) =s\left( t \right) -s\left( 0 \right) 
		-\ \ \ \ \ \ \ \ \ \ \ \ \ \ \ \ \ \ \ \ \ \ \ \ \ \ \ \ \ \ \ \ \ \ \ \ \ \ \ \ \ \ \ \ \ \ \ \ \ \\
		\int_0^t{\left(\dot{e}+\frac{1}{m\kappa_{1}^{m}}\left[ \dot{e}+\kappa_2\left[ e \right] ^n \right] ^{\frac{1}{m}-1}\left( \ddot{e}+\kappa_2n\left[ e \right] ^{n-1} \right) \dot{e} \right)dt}
	\end{split}
\end{equation*}
\end{small}

Then, the compensating controller $u_{s}$ is designed as:
\begin{small}
\begin{equation*}
	u_{s}=\varXi ^{-1}\left( -\left( \rho +\varepsilon \right) sign\left( \sigma \right) -\kappa_3\left[ \sigma \right] ^p-\kappa_4\left[ \sigma \right] ^q \right) 
\end{equation*}
\end{small}
where $\rho$ and $\varepsilon$ are small positive constant. $p$ and $q$ are constants satisfying $0<p<1$ and $q>1$. After getting nominal controller $u_0$ and compensating controller $u_{s}$, the FxTISMC is derived by using $u_c=u_0+u_s$. The stability and convergence of this controller have been proven in our previous work \cite{RN488}. 

%%%%%%%%%%%%%%%%%%%%%%%%%%%%%%%%%%%%%%%%%%%%%%%%%%%%%%%%%%%%%%%%%%%%%%%%%%%%%%%%%%%%%%%%%%%%%%%%%%%%%%%%%%%%%%
% Numerical Example
%%%%%%%%%%%%%%%%%%%%%%%%%%%%%%%%%%%%%%%%%%%%%%%%%%%%%%%%%%%%%%%%%%%%%%%%%%%%%%%%%%%%%%%%%%%%%%%%%%%%%%%%%%%%%%
\section{Numerical Example}\label{sec5}
In this section, a two-link planar robot manipulator is employed as a use case to conduct the simulation. The approach can in principle be extended to a robot with arbitrary degrees of freedom. 

\subsection{Simulation details}
The dynamics of the two-link planar robot are described as \cite{RN1987}:
\begin{small}
\begin{equation*}
	\begin{aligned}
		\tau _1&=m_2l_{2}^{2}\left( \ddot{q}_1+\ddot{q}_2 \right) +m_2l_1l_2c_2\left( 2\ddot{q}_1+\ddot{q}_2 \right) +\left( m_1+m_2 \right) l_{1}^{2}\ddot{q}_1-\\
		&m_2l_1l_2s_2\dot{q}_{2}^{2}-2m_2l_1l_2s_2\dot{q}_1\dot{q}_2+m_2l_2gc_{12}+\left( m_1+m_2 \right) l_1gc_1\\
		\tau _2&=m_2l_1l_2c_2\ddot{q}_1+m_2l_1l_2s_2\dot{q}_{1}^{2}+m_2l_2gc_{12}+m_2l_{2}^{2}\left( \ddot{q}_1+\ddot{q}_2 \right) 
	\end{aligned}
\end{equation*}
\end{small}
where $c_i=\cos \left( q_i \right) $, $c_{ij}=\cos \left( q_i+q_j \right) $, $s_i=\sin \left( q_i \right) $, and $s_{ij}=\sin \left( q_i+q_j \right)$, $i,j=1,2$. The $M\left( q \right)$, $C\left( q,\dot{q} \right)$, $G\left( q \right)$, $F\left( q \right)$, and the Jacobian of the robot are given as: 
\begin{small}
\begin{equation*}
	\begin{split}
	M\left( q \right) =\left[ \begin{matrix}
	m_2l_{2}^{2}+2m_2l_1l_2c_2+\left( m_1+m_2 \right) l_{1}^{2}&		m_2l_{2}^{2}+m_2l_1l_2c_2\\
	m_2l_{2}^{2}+m_2l_1l_2c_2&		m_2l_{2}^{2}\\
\end{matrix} \right] \ \ \ \ \ \ \ \ \ \ \ \ \ \ \ \ \\
 C\left( q,\dot{q} \right) =\left[ \begin{array}{c}
	-m_2l_1l_2s_2\dot{q}_{2}^{2}-2m_2l_1l_2s_2\dot{q}_1\dot{q}_2\\
	m_2l_1l_2s_2\dot{q}_{1}^{2}\\
\end{array} \right] \ \ \ \ \ \ \ \ \ \ \ \ \ \ \ \ \ \ \ \ \ \ \ \ \ \ \ \\
	G\left( q \right) =\left[ \begin{array}{c}
	m_2l_2gc_{12}+\left( m_1+m_2 \right) l_1gc_1\\
	m_2l_2gc_{12}\\
\end{array} \right]\ \ \ \ \ \ \ \ \ \ \ \ \ \ \ \ \ \ \ \ \ \ \ \ \ \ \ \ \\
 F\left( q \right) =\left[ \begin{array}{c}
	2c_1s_2+5c_{1}^{2}\\
	-2c_1s_2-5c_{1}^{2}\\
\end{array} \right], 
J\left( q \right) =\left[ \begin{matrix}
	-l_1s_1-l_2s_{12}&		-l_2s_{12}\\
	l_1c_1+l_2c_{12}&		l_2c_{12}\\
\end{matrix} \right] \ \ \ \ \ \ \ \ \ \ \ \ \ \ \ \ \ \ \
	\end{split}
\end{equation*}
\end{small}
\begin{table}[t]
	\centering
	\caption{Simulation parameters.}
	\begin{tabular}{cc}  
		\hline
		Modules & Parameters\\  
		\hline
		Initial value &  \makecell{$q\left(0\right)=\left[0.5236,2.0944\right]^T$ \\ $x\left(0\right)=\left[0,0\right]^T$ } \\ 
		\hline
		Robot dynamics & \makecell{$m_1=1.5kg$, $m_2=1.0kg$ \\$l_1=l_2=0.3m$} \\ 
		\hline
		ECBFs-QP &  \makecell{$K_{\max_i}=\left[500,50\right]$ \\ $K_{\min_i}=\left[500,50\right]$ \\$K=\left[700,70\right]$}\\ 
		\hline
		Admittance control &   \makecell{$k_{m_i}=20$, $k_{b_i}=20$, $k_{k_i}=100$ \\ $a_1=1$, $a_2=2$} \\ 
		\hline
		Low-Level controller &  \makecell{$\lambda_1=3$, $\lambda_2=20$, $\lambda_3=50$ \\ $\alpha = m=p=\frac{5}{7}$, $\beta  = n=q=\frac{5}{3}$ \\$\kappa_1 = \kappa_3=20$, $\kappa_2 = \kappa_4=50$}  \\ 
		\hline
		Safety constraints & \makecell{workspace boundary $\left[-0.13,0.13\right]$ \\ obstacle position $\left[-0.07,0.07\right]$ \\safe distance $r=0.04$} \\ 
		\hline 
	\end{tabular}
\end{table}

\begin{figure}[htbp]
	\begin{subfigure}{0.29\textwidth}
		\centering
		\includegraphics[width=.99\linewidth]{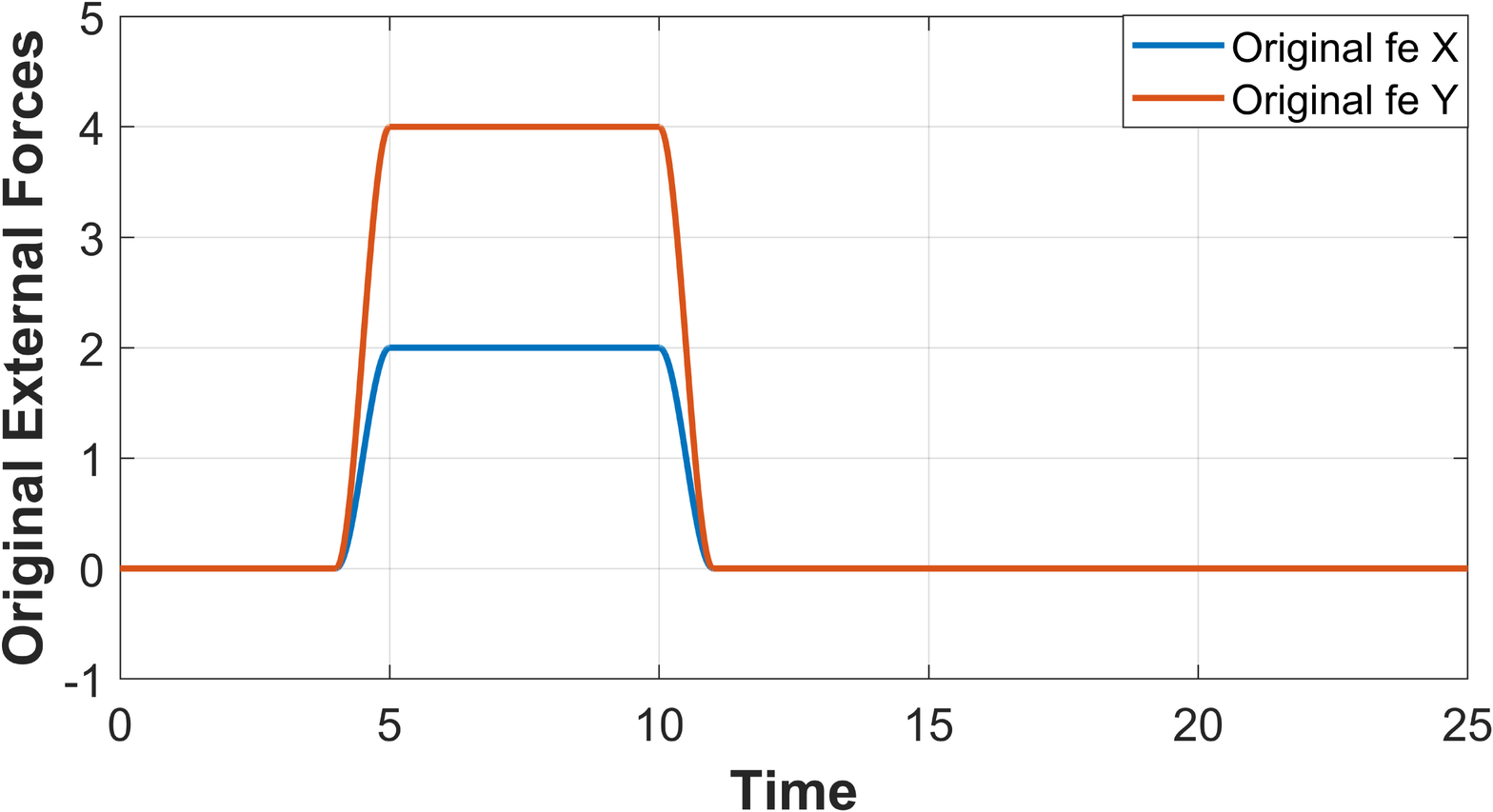}
		\caption{Time-varying external human forces.}
		\label{f1a}
	\end{subfigure}%
	\begin{subfigure}{0.19\textwidth}
		\centering
		\includegraphics[width=.99\linewidth]{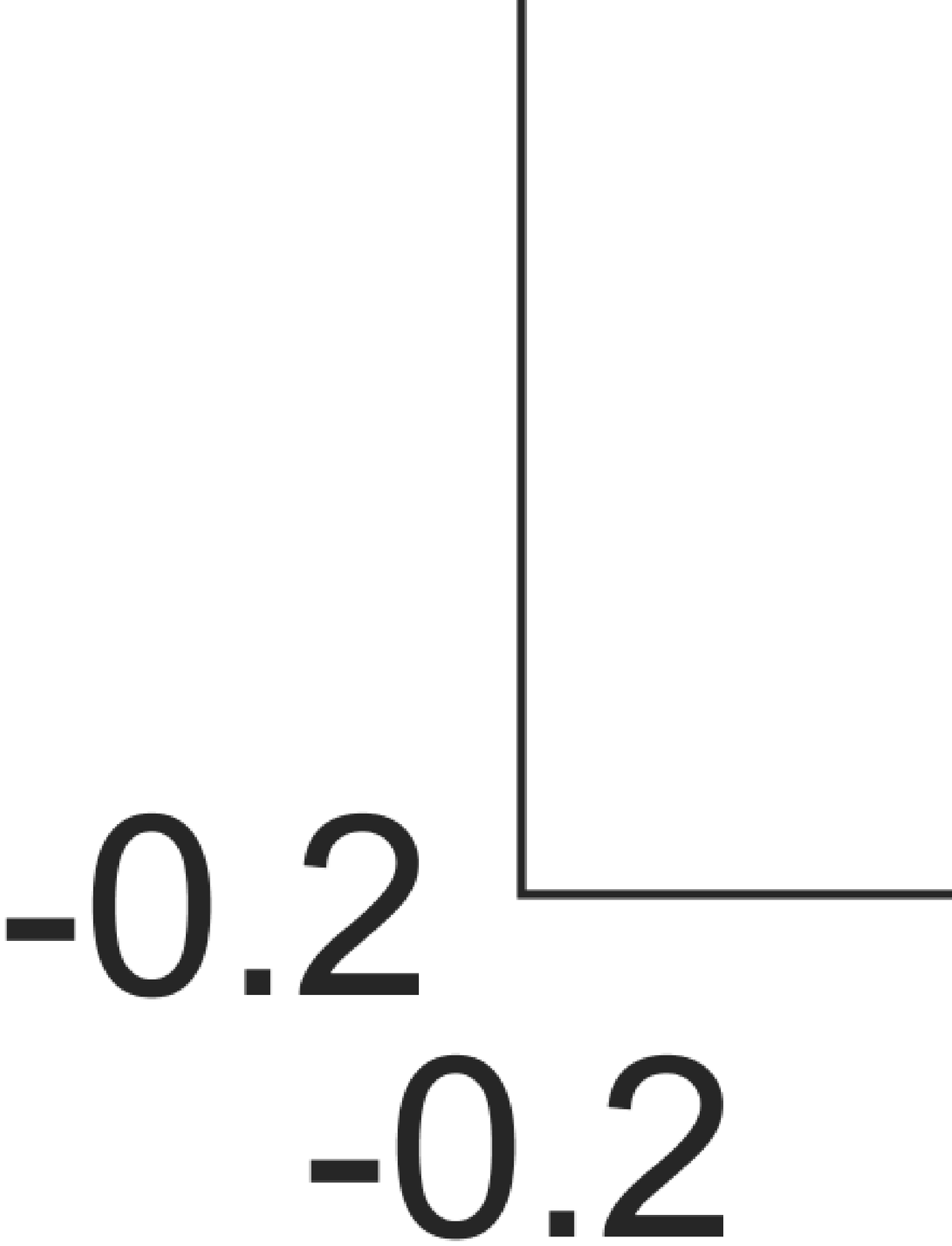}
		\caption{Unsafe trajectory.}
		\label{f1b}
	\end{subfigure}
	\caption{Traditional admittance control.}
	\label{f1}
\end{figure}

Assume the desired trajectory of the end-effector is a circle with the origin as the centre and a radius of 0.14 (as can be seen in Fig. (\ref{f1b}), which can be formulated as \cite{RN100}:
\begin{small}
\begin{equation}
	\label{eq11}
	\begin{aligned}
		x_d\left( t \right) &= 0.14\cos \left( 0.5t \right) \\
		y_d\left( t \right) &=0.14\sin \left( 0.5t \right) 	
	\end{aligned}
\end{equation}
\end{small}

The external human forces are given by \cite{RN100}:
\begin{small}
\begin{equation}
	\label{eq36}
	f_{e_i}\left( t \right) =\left\{ \begin{array}{l}
		0\ \ \ \ \ \ \ \ \ \ \ \ \ \ \ \ \ \ \ \ t<5\ or\ t\ge 11\\
		a_i\left( 1-\cos \pi t \right) \ \ \ \ \ \ 4\le t<5\\
		2a_i\ \ \ \ \ \ \ \ \ \ \ \ \ \ \ \ \ \ \ \  5\le t<10\\
		a_i\left( 1+\cos \pi t \right) \ \ \ \ \ \ 10\le t<11
	\end{array} \right. 
\end{equation}
\end{small}

The external human forces are applied when $t=4s$ and removed at $t=11s$, as depicted in Fig. \ref{f1a}. By using the traditional admittance control, as can be seen in Fig. \ref{f1b}, the external human forces cause the deformation of the trajectory, and the resulting trajectory crosses the boundaries of the workspace (i.e., red dotted line in Fig. \ref{f1b}). In the following subsection, we integrate our proposed ECBFs-QP based adaptive admittance control into our control framework so that the safety constraints can be strictly guaranteed, while allowing the robot to comply with human interaction safely inside the task space with natural fluidity. The parameters of the simulation are given in Table. 1.
\subsection{Simulation results}
\begin{figure}[htbp]
	\begin{subfigure}{0.23\textwidth}
		\centering
		\includegraphics[width=.99\linewidth]{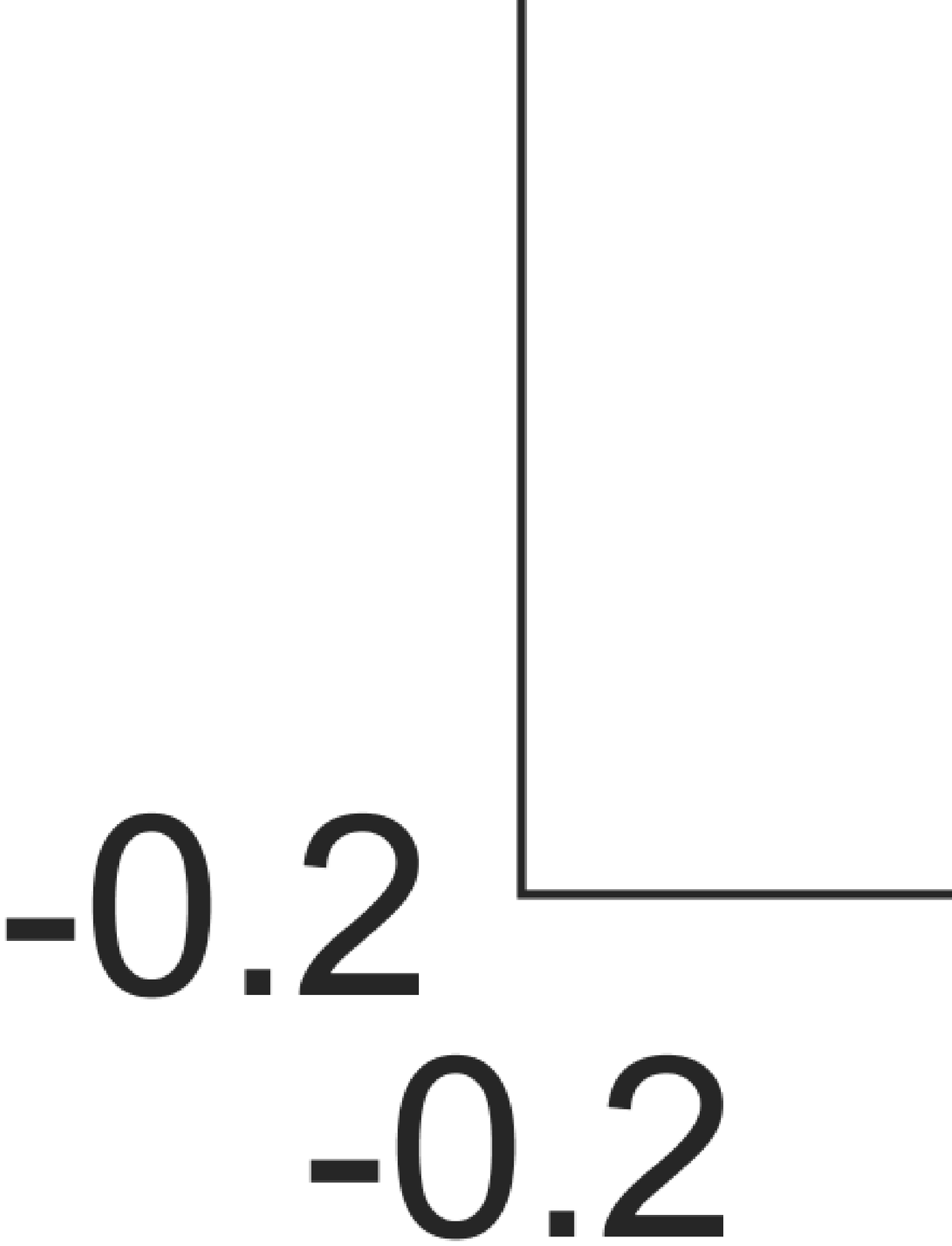}
		\caption{2D trajectory.}
		\label{f3a}
	\end{subfigure}%
	\begin{subfigure}{0.23\textwidth}
		\centering
		\includegraphics[width=.99\linewidth]{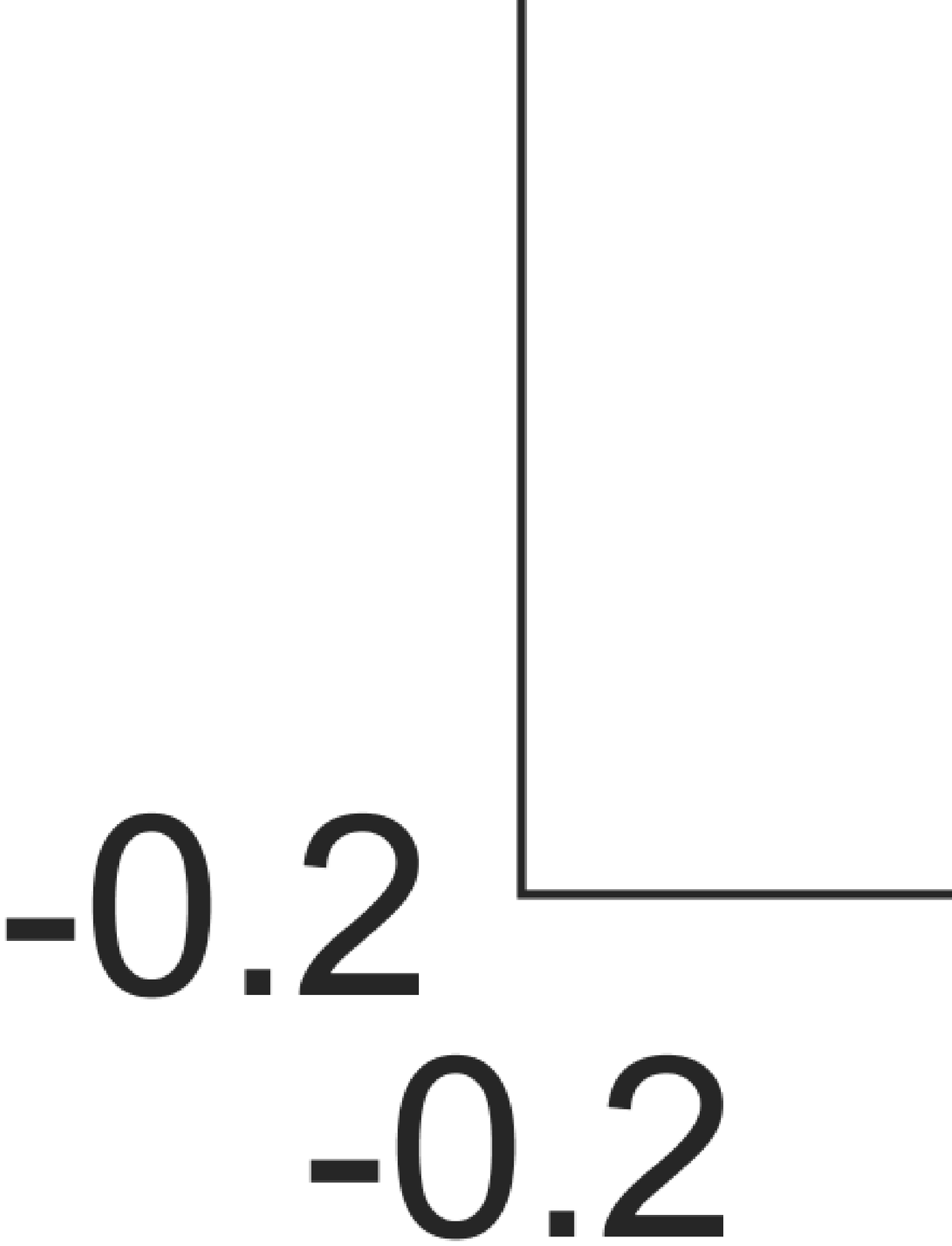}
		\caption{2D trajectory.}
		\label{f3b}
	\end{subfigure}\\
	\begin{subfigure}{0.24\textwidth}
		\centering
		\includegraphics[width=.99\linewidth]{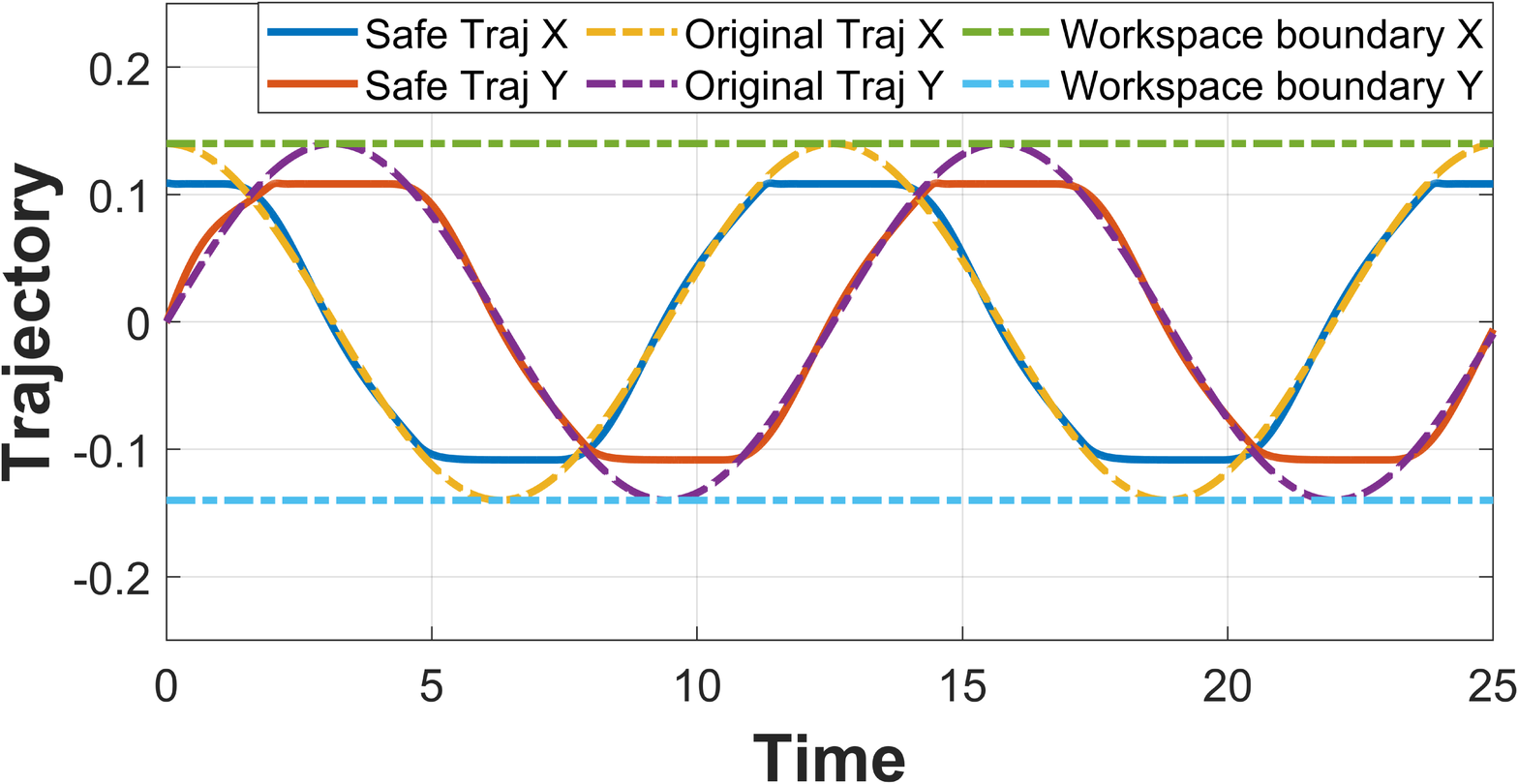}
		\caption{Trajectory on X and Y axis.}
		\label{f3c}
	\end{subfigure}
	\begin{subfigure}{0.24\textwidth}
		\centering
		\includegraphics[width=.99\linewidth]{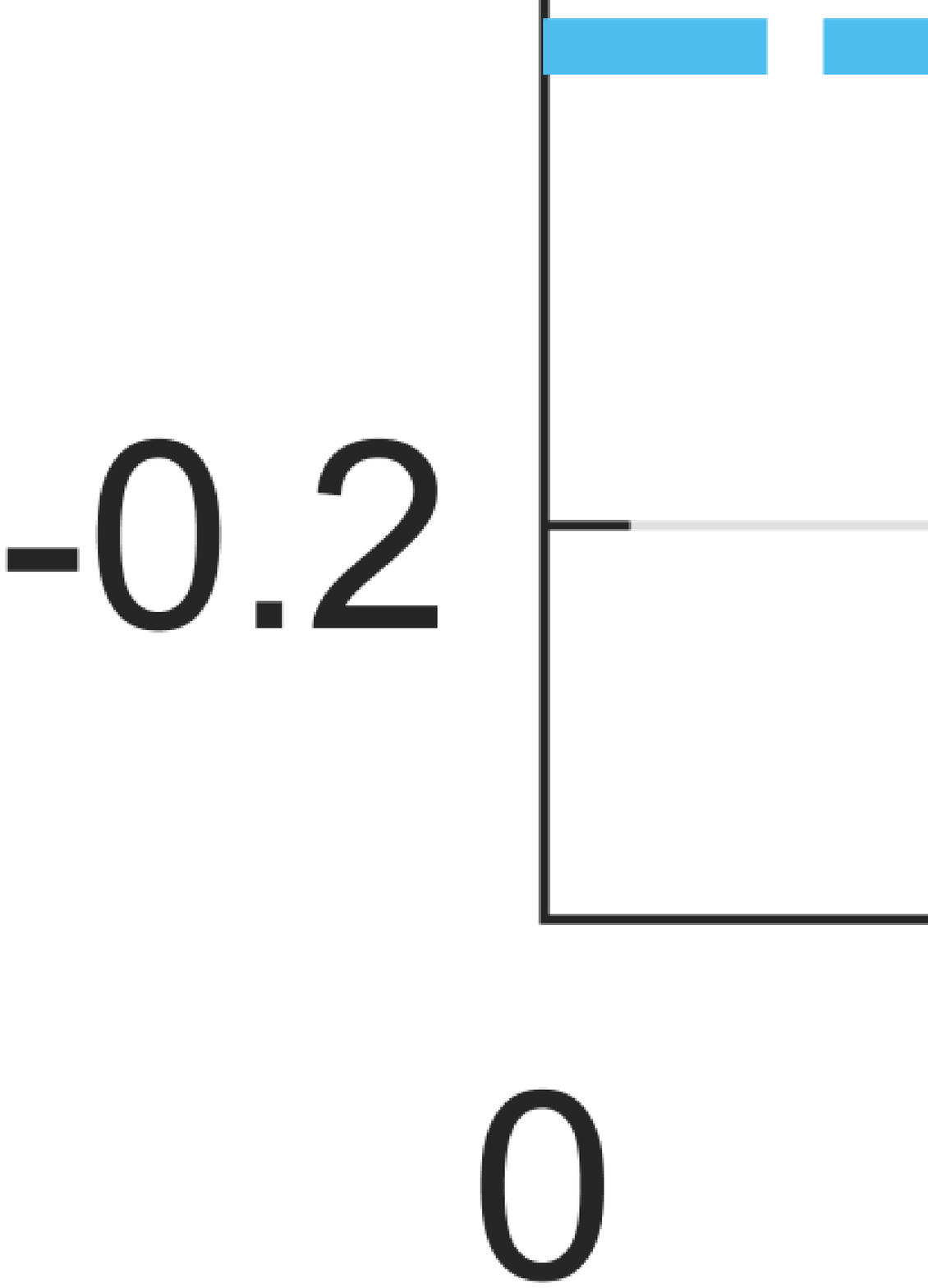}
		\caption{Trajectory on X and Y axis.}
		\label{f3d}
	\end{subfigure}\\
	\begin{subfigure}{0.24\textwidth}
		\centering
		\includegraphics[width=.99\linewidth]{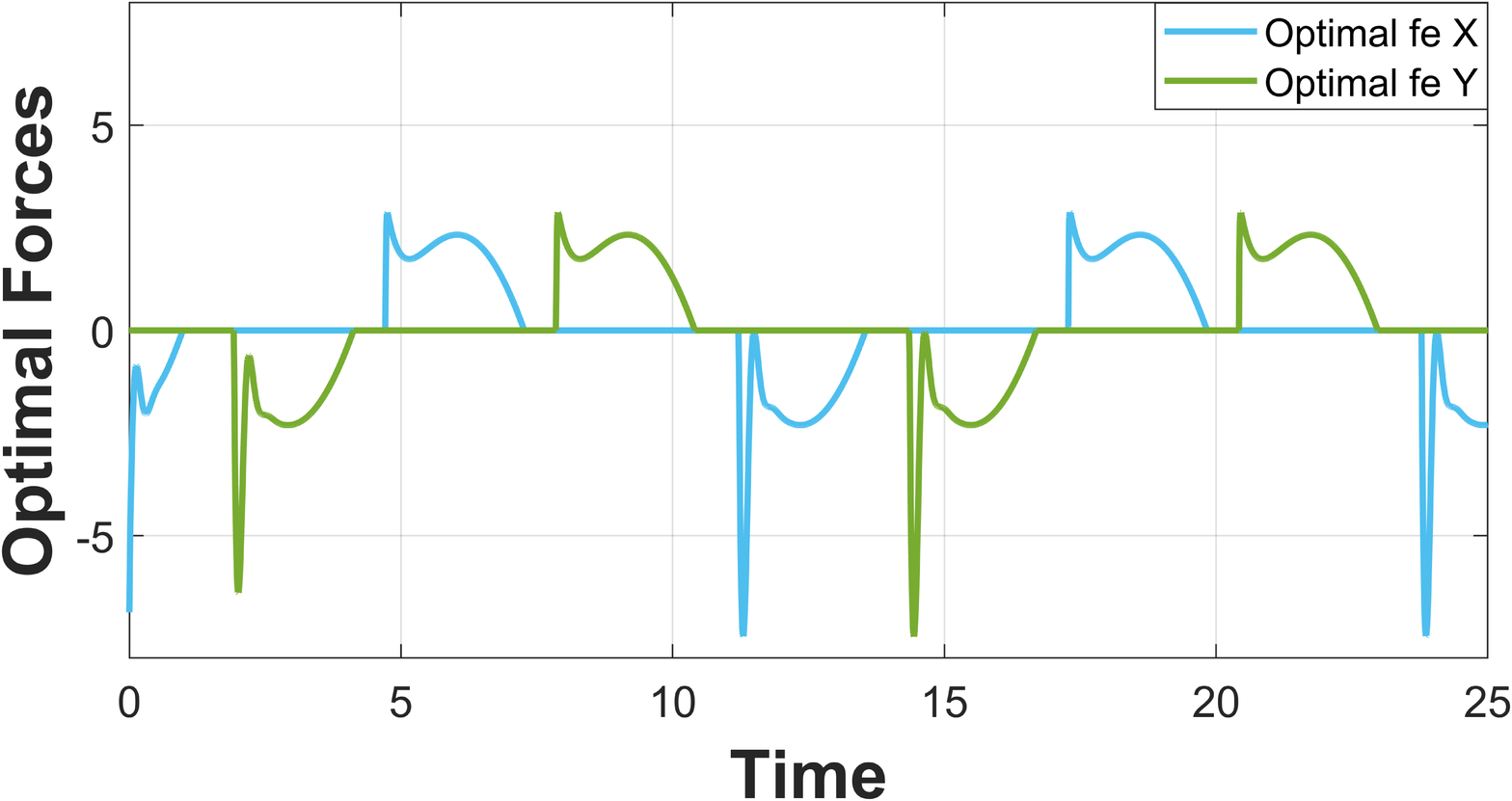}
		\caption{Interaction forces.}
		\label{f3e}
	\end{subfigure}
	\begin{subfigure}{0.24\textwidth}
		\centering
		\includegraphics[width=.99\linewidth]{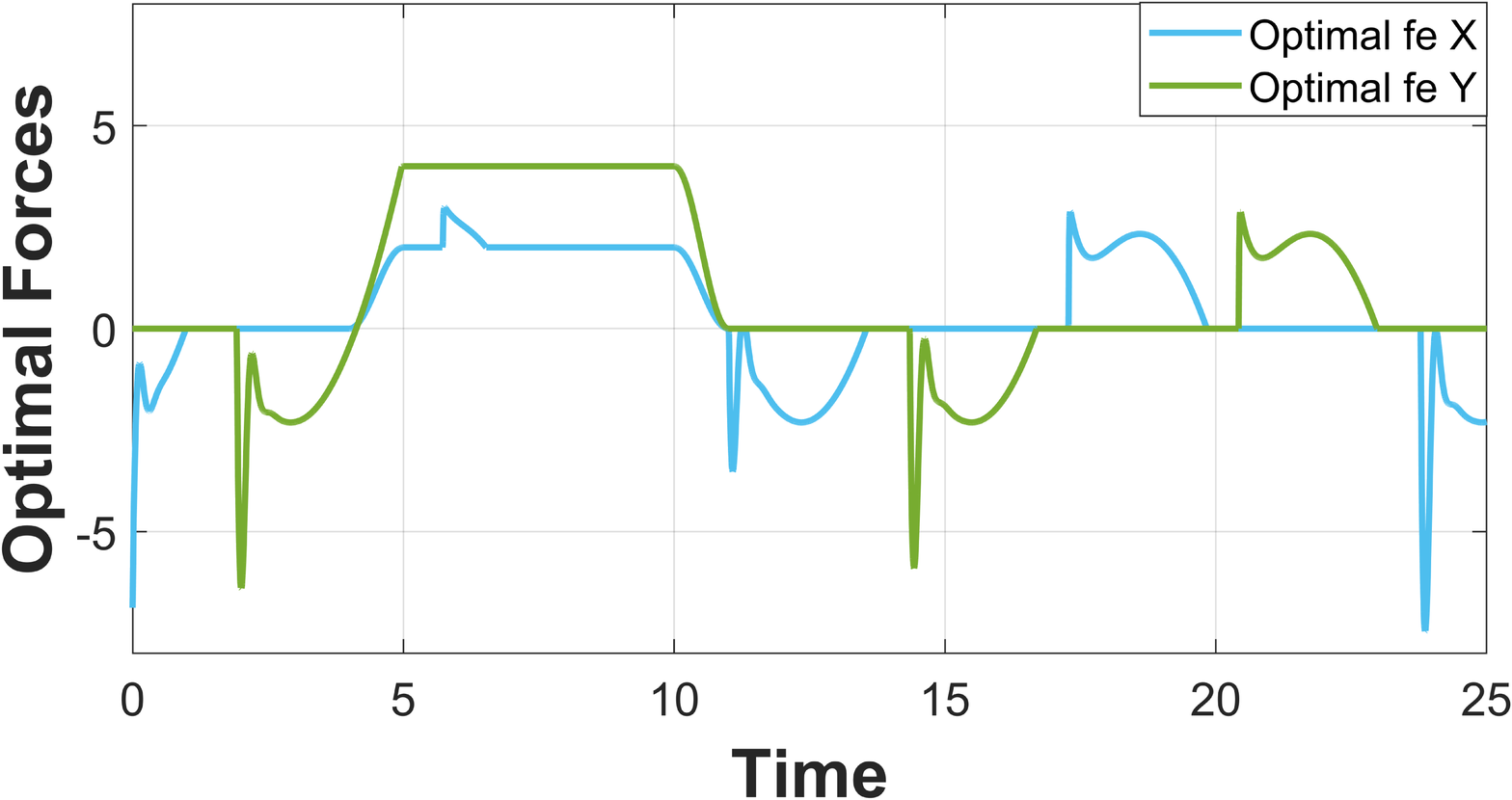}
		\caption{Interaction forces.}
		\label{f3f}
	\end{subfigure}
	\caption{Robot manipulator with workspace constraints}
	\label{f3}
\end{figure}
In the case of workspace constraints, we illustrate the effectiveness of our proposed approach by applying and removing external forces. As can be seen in Fig. \ref{f3a}, the original desired trajectory $x_d$ is an ideal circle (indicated with the solid blue line). When we apply workspace constraints (i.e., rectangle indicated with the red dotted lines), the original desired trajectory is modified to the safe trajectory that is constrained within the workspace and keeps a safe distance from the boundary. When we apply the external human forces (\ref{eq36}) to both links of the robot manipulator (as shown in Fig. \ref{f3b}), the robot's trajectory changes in response to external human forces, but still remains within the restricted area. Fig. \ref{f3c} and Fig. \ref{f3d} show the Cartesian trajectory of the end-effector on the X-axis and Y-axis, separately. When $4<t<11$, compared with Fig. \ref{f3c}, the trajectory in Fig. \ref{f3d} is deformed in order to comply with the external forces. It is clear that the external loop based on the admittance control ensures the compliance of robot motion, and the internal feedback loop based on ECBFs-QP ensures the safety constraints. Fig. \ref{f3e} and Fig. \ref{f3f} show the optimal interaction forces derived by the ECBFs-QP. It is clear that our methods can provide effective compensative force feedback to admittance control such that both compliance and safety can be guaranteed.
\begin{figure}[htbp]
	\begin{subfigure}{0.23\textwidth}
		\centering
		\includegraphics[width=.99\linewidth]{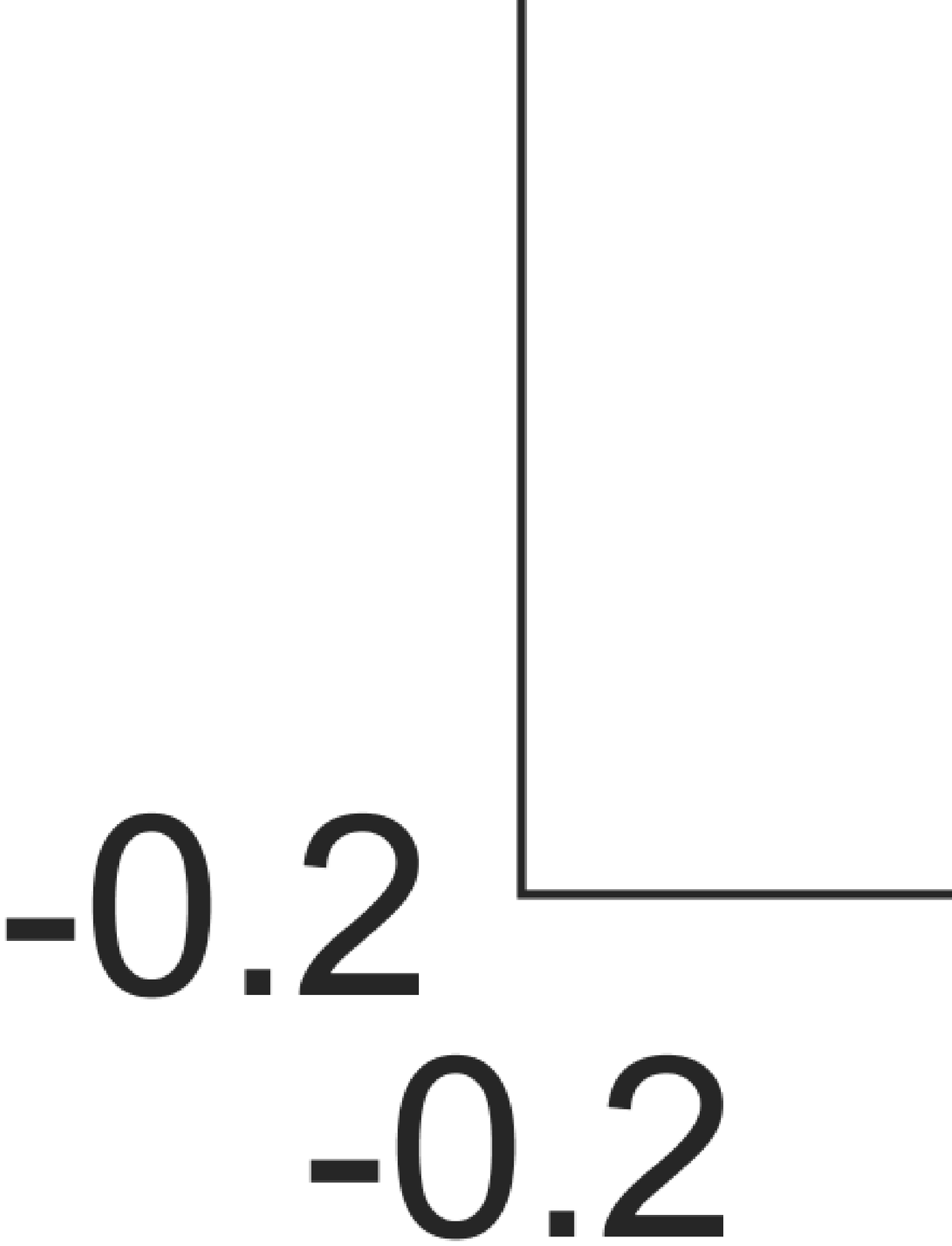}
		\caption{2D trajectory.}
		\label{f4a}
	\end{subfigure}%
	\begin{subfigure}{0.23\textwidth}
		\centering
		\includegraphics[width=.99\linewidth]{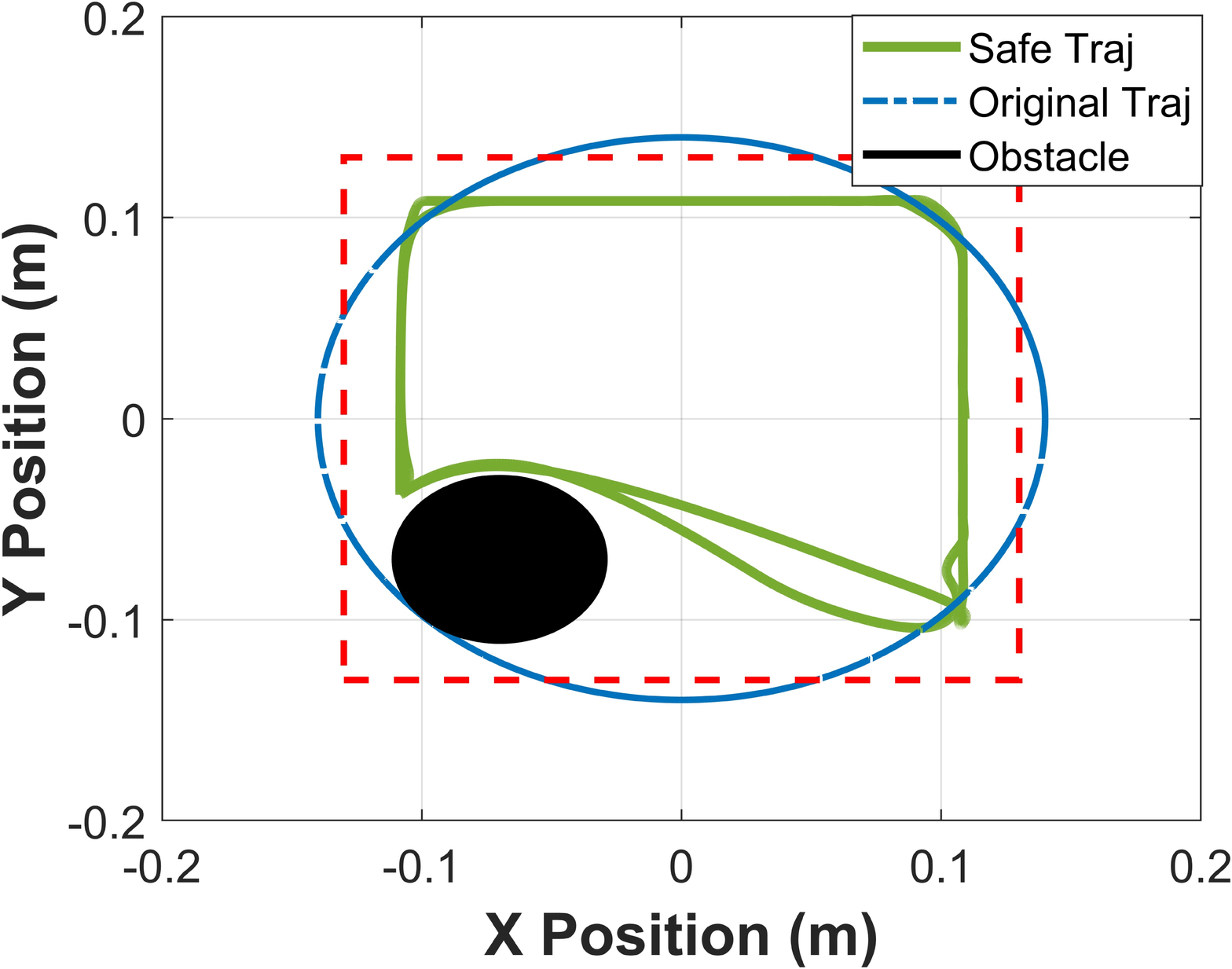}
		\caption{2D trajectory.}
		\label{f4b}
	\end{subfigure}\\
	\begin{subfigure}{0.24\textwidth}
		\centering
		\includegraphics[width=.99\linewidth]{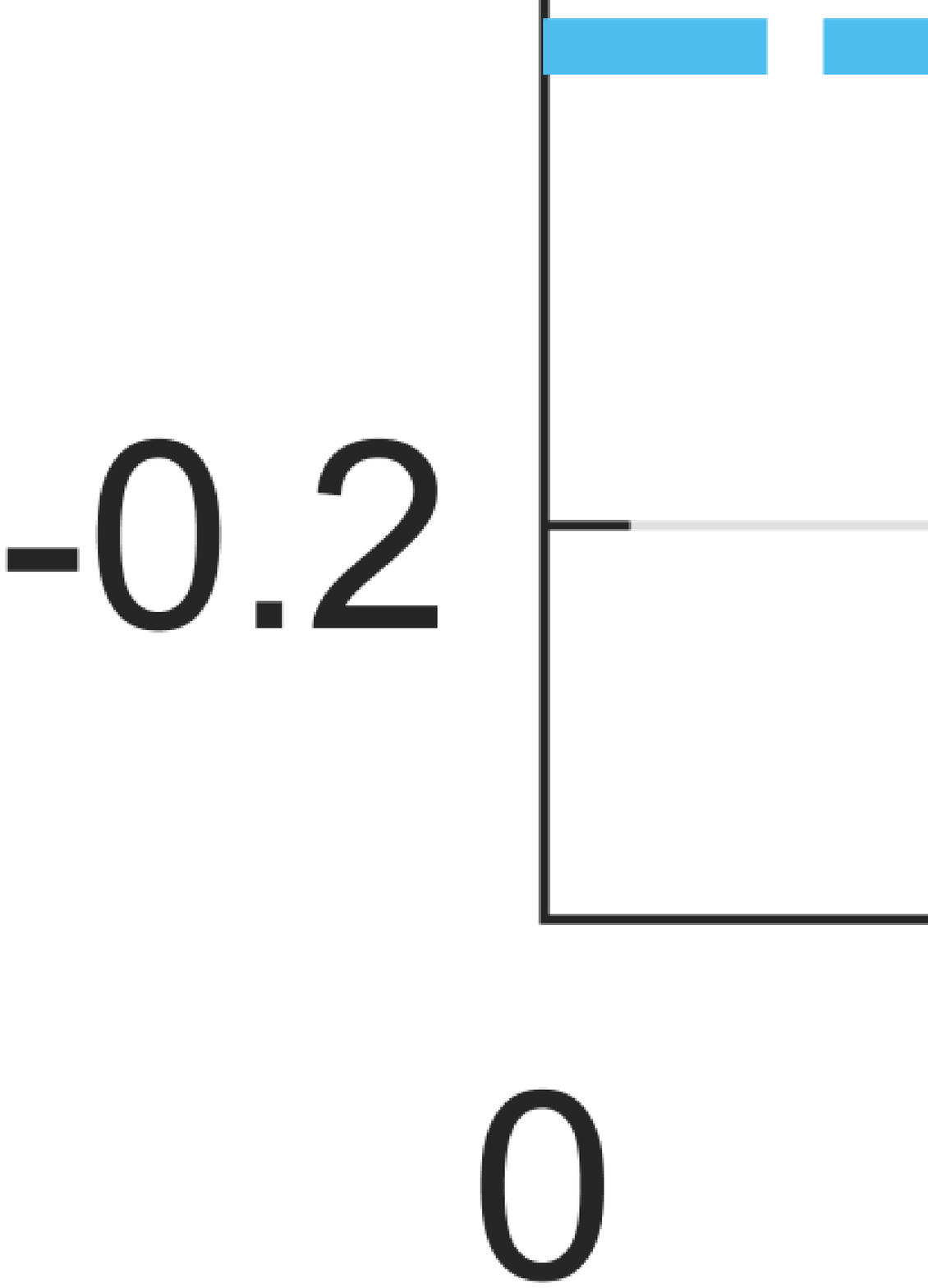}
		\caption{Trajectory on X and Y axis.}
		\label{f4c}
	\end{subfigure}
	\begin{subfigure}{0.24\textwidth}
		\centering
		\includegraphics[width=.99\linewidth]{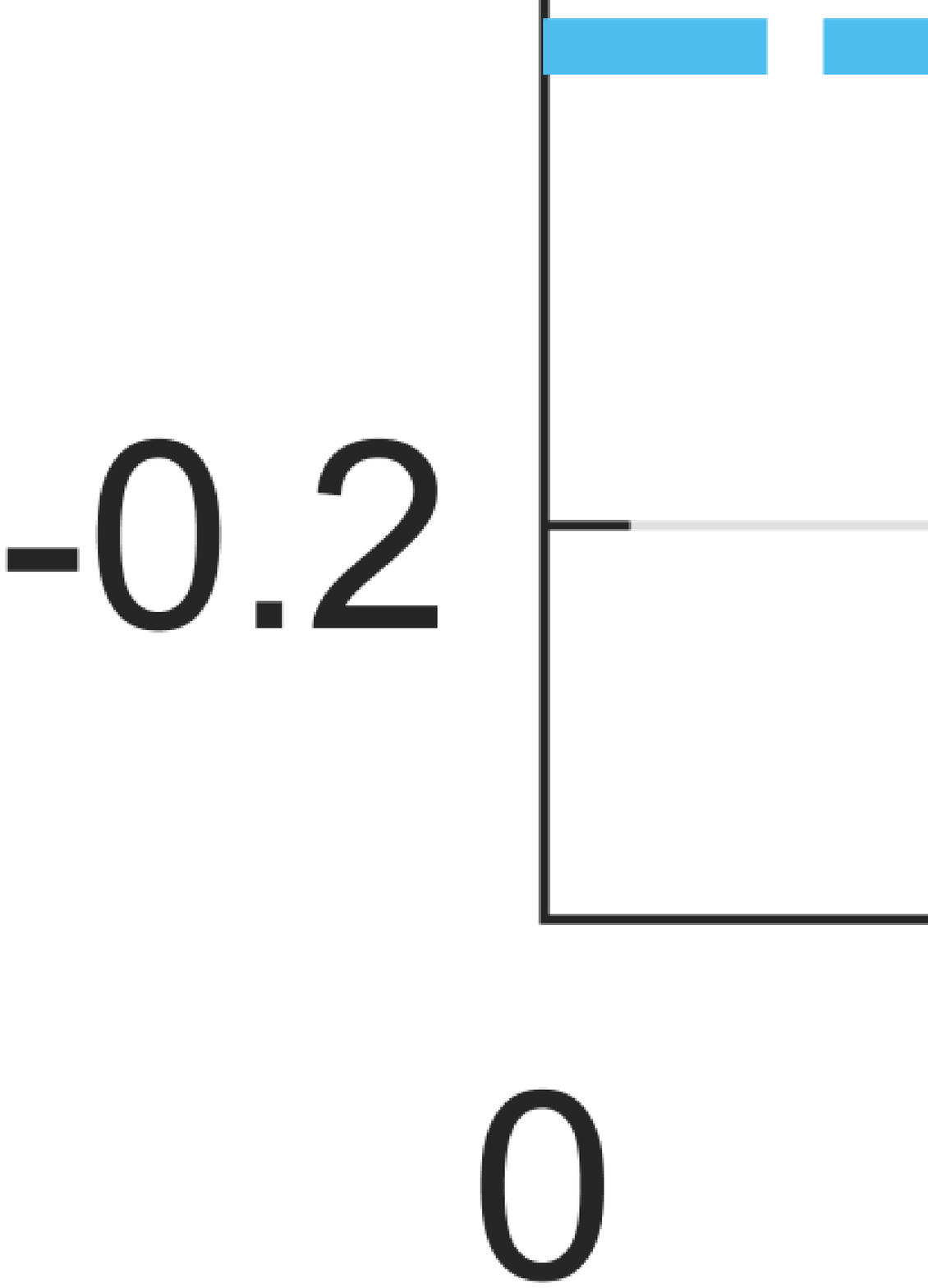}
		\caption{Trajectory on X and Y axis.}
		\label{f4d}
	\end{subfigure}\\
	\begin{subfigure}{0.24\textwidth}
		\centering
		\includegraphics[width=.99\linewidth]{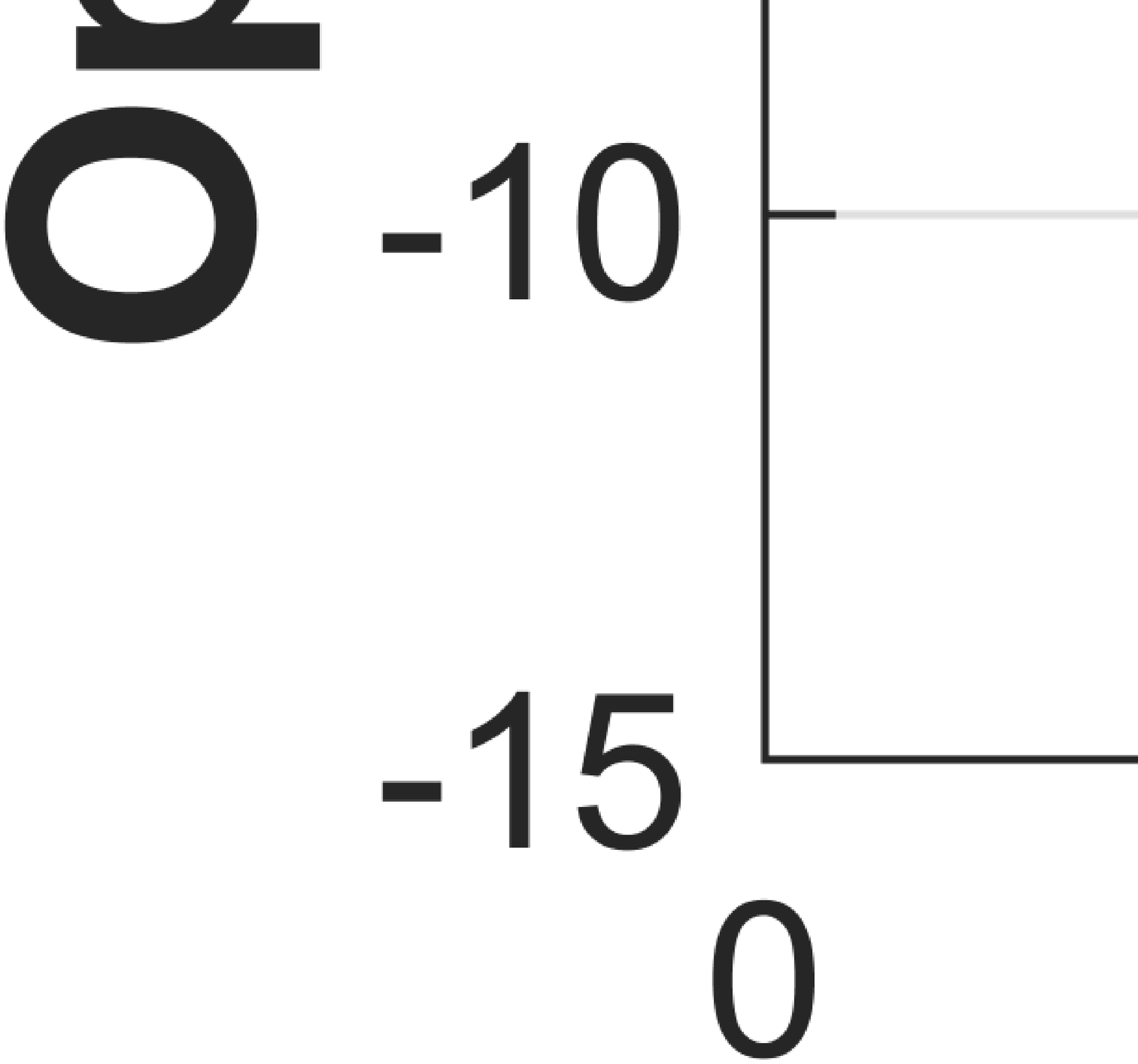}
		\caption{Interaction forces.}
		\label{f4e}
	\end{subfigure}
	\begin{subfigure}{0.24\textwidth}
		\centering
		\includegraphics[width=.99\linewidth]{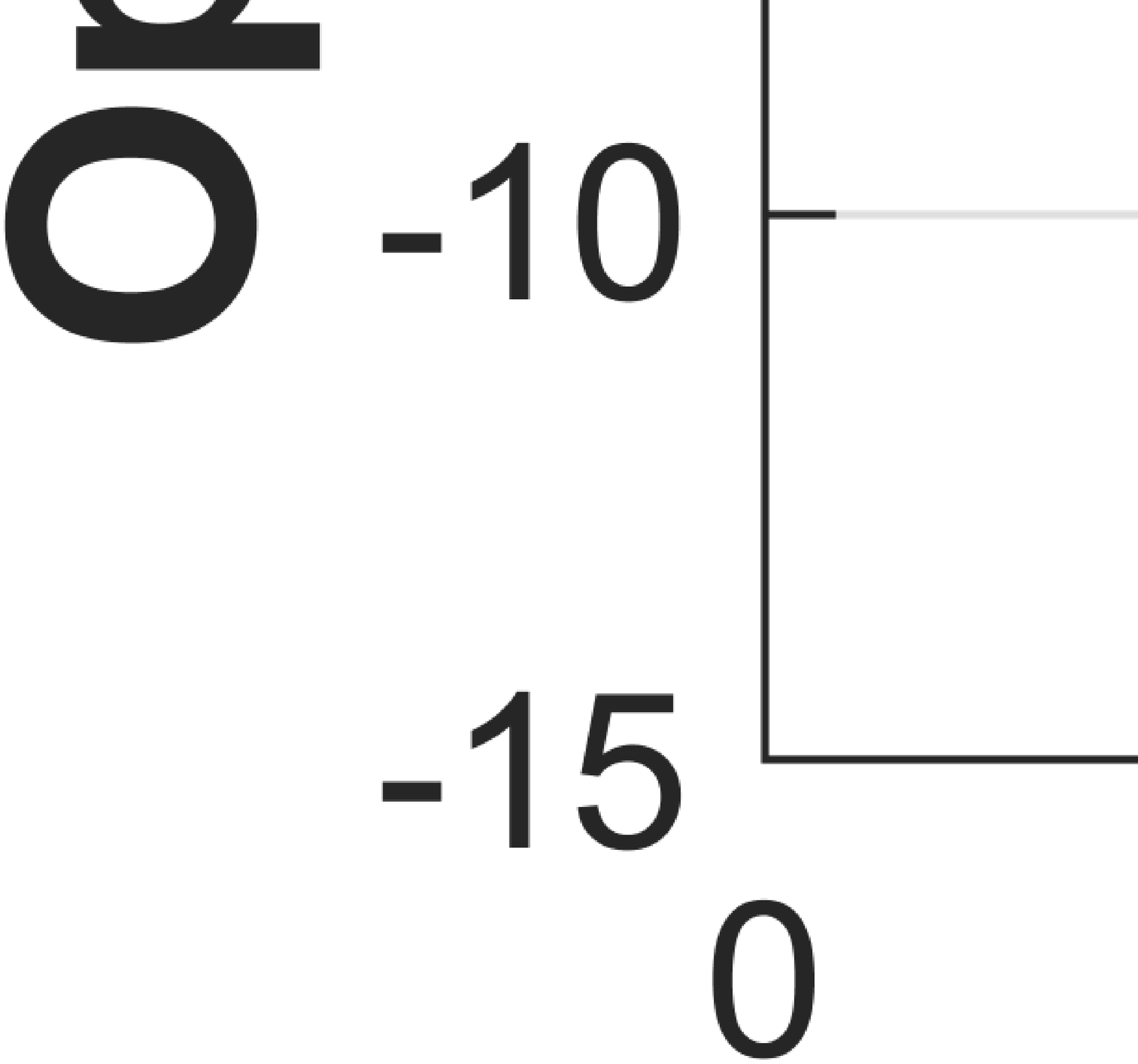}
		\caption{Interaction forces.}
		\label{f4f}
	\end{subfigure}
	\caption{Robot manipulator with both two constraints}
	\label{f4}
\end{figure}

In the case of obstacle avoidance, we firstly achieve obstacle avoidance only, and then synthesize the workspace constraints and obstacle constraints together in our proposed framework. As can be seen in Fig. \ref{f4a}, when we only consider obstacle avoidance, the robot moves into an unsafe proximity of the workspace boundary when trying to avoid the collision. In Fig. \ref{f4b}, after integrating the workspace constraints into the system, the robot avoids all hazards. The Cartesian trajectories of the end-effector on the X-axis and Y-axis are shown in Fig. \ref{f4c} and Fig. \ref{f4d}, and the optimal interaction forces derived from the ECBFs-QP are shown in Fig. \ref{f4e} and Fig. \ref{f4f}. It is clear that in the presence of human external forces, both compliance and safety can be ensured by our proposed method.

\section{Conclusions}\label{sec6}
In this letter, a novel control framework based on admittance control, exponential control barrier functions and the quadratic program is proposed to achieve both compliance and safety for human-robot interaction.
In particular, a virtual force feedback for admittance control is provided in real-time by using the ECBFs-QP framework as a compensator of the external human forces. Therefore, the safety of the proposed robot control system has higher robustness for external force disturbances, while simultaneously providing human-friendly dynamic behaviour. The simulation results show that the proposed approach can enforce both safety and compliance. In future work, constraints will be considered for each joint (not only the end-effector). In addition, the barrier Lyapunov function will be discussed and integrated into the system to further enforce safety.

\addtolength{\textheight}{-4cm}   % This command serves to balance the column lengths

%%%%%%%%%%%%%%%%%%%%%%%%%%%%%%%%%%%%%%%%%%%%%%%%%%%%%%%%%%%%%%%%%%%%%%%%%%%%%%%%

%%%%%%%%%%%%%%%%%%%%%%%%%%%%%%%%%%%%%%%%%%%%%%%%%%%%%%%%%%%%%%%%%%%%%%%%%%%%%%%%

%%%%%%%%%%%%%%%%%%%%%%%%%%%%%%%%%%%%%%%%%%%%%%%%%%%%%%%%%%%%%%%%%%%%%%%%%%%%%%%%
%\section*{ACKNOWLEDGMENT}

%%%%%%%%%%%%%%%%%%%%%%%%%%%%%%%%%%%%%%%%%%%%%%%%%%%%%%%%%%%%%%%%%%%%%%%%%%%%%%%%
%\input{references.tex}
\bibliographystyle{IEEEtran}
\bibliography{references}

\end{document}